\documentclass[%
 reprint,
superscriptaddress,
 amsmath,amssymb,
 aps,
]{revtex4-2}

\usepackage{graphicx}
\usepackage{dcolumn}
\usepackage{bm}
\usepackage{algorithm}
\usepackage{algpseudocode} 
\usepackage{bbm}
\usepackage{tikz}
\usetikzlibrary{arrows.meta, positioning, fit}
\usepackage{booktabs}

\begin{document}

\preprint{APS/123-QED}

\title{ADAPT: Lightweight, Long-Range Machine Learning Force Fields Without Graphs}

\author{Evan Dramko}
\affiliation{Department of Computer Science, Rice University, Houston, TX, USA}

\author{Yizhi Zhu}
\affiliation{Department of Materials Science and Nanoengineering, Rice University, Houston, TX, USA}
\affiliation{Rice Advanced Materials Institute, Rice University, Houston, TX, USA}
\affiliation{Thayer School of Engineering, Dartmouth College, Hanover, NH, USA}

\author{Yihuang Xiong}
\affiliation{Thayer School of Engineering, Dartmouth College, Hanover, NH, USA}

\author{Geoffroy Hautier}
\email{geoffroy.hautier@rice.edu} 
\affiliation{Department of Materials Science and Nanoengineering, Rice University, Houston, TX, USA}
\affiliation{Rice Advanced Materials Institute, Rice University, Houston, TX, USA}
\affiliation{Thayer School of Engineering, Dartmouth College, Hanover, NH, USA}

\author{Thomas Reps}
\affiliation{Department of Computer Sciences, University of Wisconsin--Madison, Madison, WI, USA}

\author{Christopher Jermaine}
\affiliation{Department of Computer Science, Rice University, Houston, TX, USA}

\author{Anastasios Kyrillidis}
\affiliation{Department of Computer Science, Rice University, Houston, TX, USA}

\date{Sept. 3, 2026. Published in \textit{Physical Review Research} \textbf{8}, 033266 (2026). 
\url{https://doi.org/10.1103/3xf3-ts5f}}

\begin{abstract}
Point defects play a central role in driving the properties of materials. First-principles methods are widely used to compute defect energetics and structures, including at scale for high-throughput defect databases. However, these methods are computationally expensive, making machine-learning force fields (MLFFs) an attractive alternative for accelerating structural relaxations. Most existing MLFFs are based on graph neural networks (GNNs), which can suffer from oversmoothing, oversquashing, and poor representation of long-range interactions. Both of these issues are especially of concern when modeling point defects. To address these challenges, we introduce the \textit{Accelerated Deep Atomic Potential Transformer} (ADAPT), an MLFF that replaces graph representations with a direct coordinates-in-space formulation and explicitly considers all pairwise atomic interactions. Atoms are treated as “tokens,” with a Transformer encoder modeling their interactions. Applied to a dataset of silicon point defects, ADAPT achieves a $\sim 22\%$ reduction in force and a $\sim 40$ reduction in energy prediction error relative to a state-of-the-art GNN-based model, while requiring only a fraction of the computational cost.
\end{abstract}

\maketitle



\section{Introduction}
\begin{figure*}[htbp]
\centering

\begin{tikzpicture}[
    >=Stealth,
    node distance=1.2cm,
mlip/.style={
    rounded corners=6pt,
    draw=cyan!60!blue,
    fill=cyan!35,
    fill opacity=1,
    text opacity=1,
    minimum width=1.3cm,
    minimum height=1.0cm,
    font=\bfseries
},
    dft/.style={
        rounded corners=6pt,
        draw=teal!60!black,
        fill=teal!35,
        fill opacity=0.5,
        text opacity=1,
        minimum width=1.3cm,
        minimum height=1.0cm,
        font=\bfseries
    },
    arr/.style={->, line width=0.8pt, gray!70},
    stagebox/.style={
        rounded corners=8pt,
        draw=black,
        dashed,
        inner sep=4pt
    }
]

\node (x0) {\includegraphics[width=2cm]{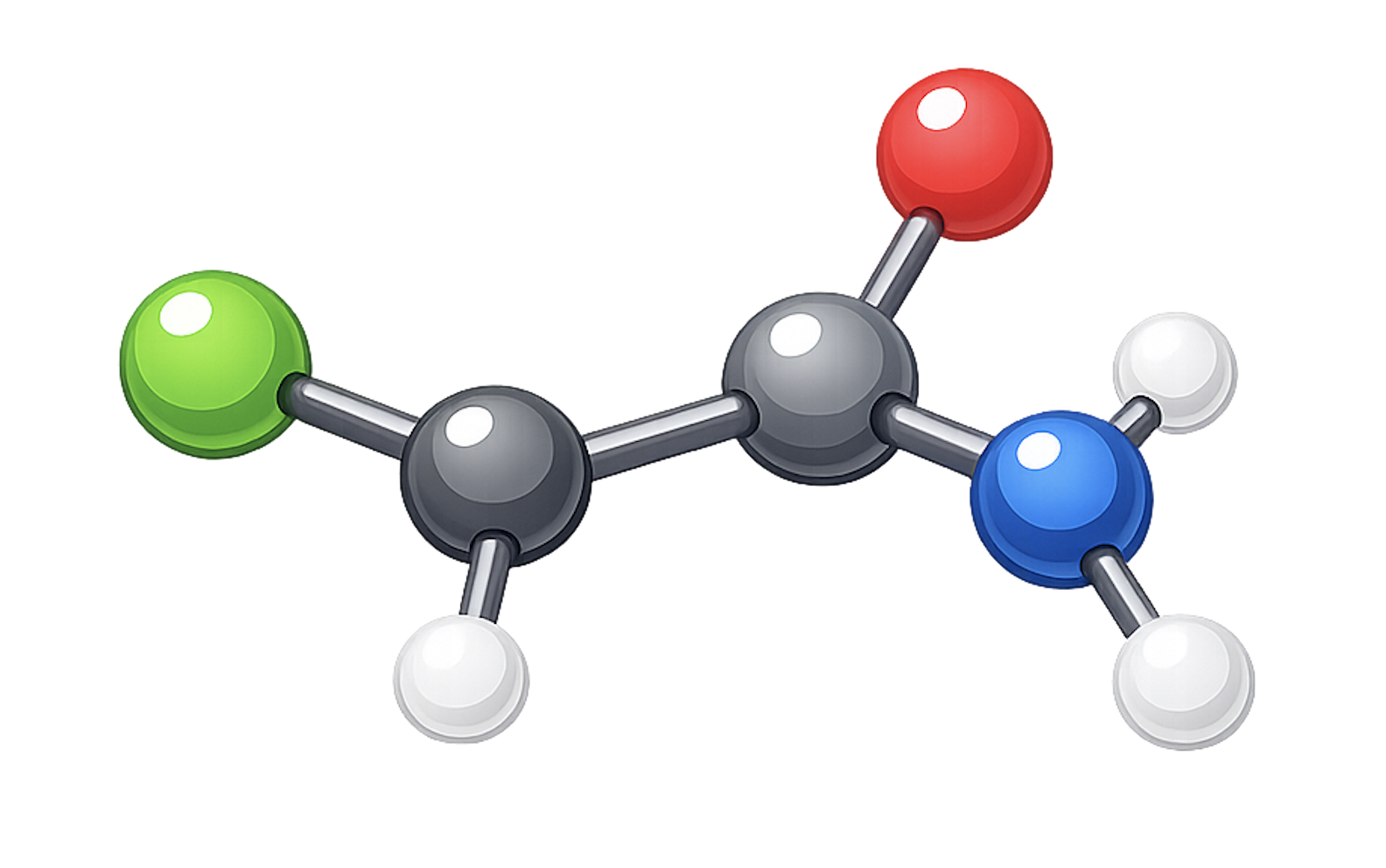}};

\node[mlip, right=of x0] (mlip1) {MLFF};

\node[right=of mlip1] (x1) {\includegraphics[width=2cm]{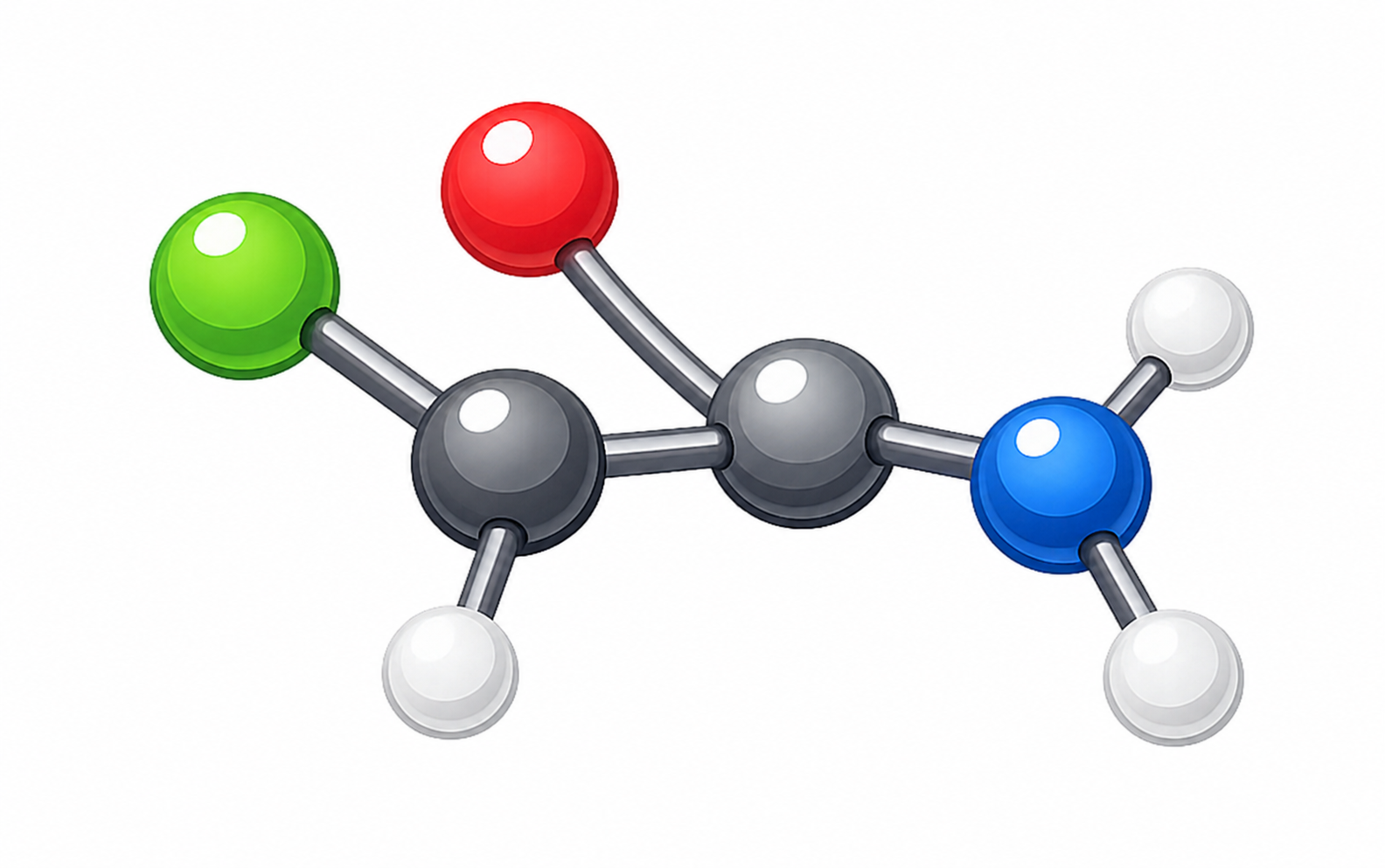}};

\node[dft, right=of x1] (dft1) {DFT};

\node[right=of dft1] (x2) {\includegraphics[width=2cm]{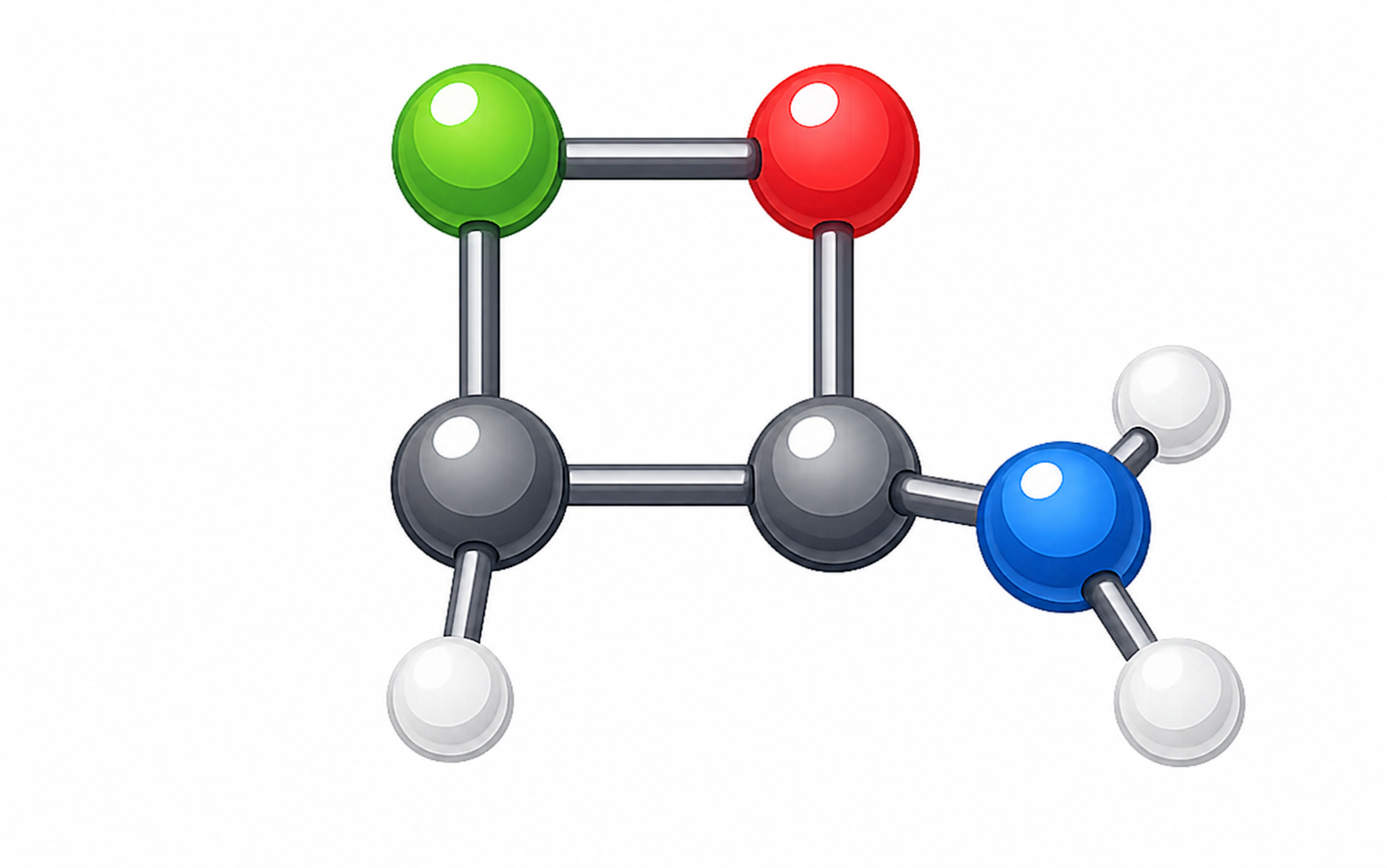}};

\node[
    stagebox,
    fit=(mlip1)
] (mlipbox) {};

\node[
    font=\bfseries\small,
    above=0.15cm of mlipbox
]
{\(\times k_{\mathrm{MLFF}}\)};

\node[
    stagebox,
    fit=(dft1)
] (dftbox) {};

\node[
    font=\bfseries\small,
    above=0.15cm of dftbox
]
{\(\times k_{\mathrm{DFT}}\)};

\draw[arr] (x0) -- (mlipbox);
\draw[arr] (mlipbox) -- (x1);
\draw[arr] (x1) -- (dftbox);
\draw[arr] (dftbox) -- (x2);

\node[below=0.35cm of mlip1, font=\scriptsize, align=center]
    {\(\mathbf{F}_{\mathrm{i}}=f_\theta(\mathbf{x}_i)\)\\[-1pt]
     \(\mathbf{x}_{i+1}=\mathbf{x}_i+\eta\mathbf{F}_{i}\)};

\node[below=0.35cm of dft1, font=\scriptsize, align=center]
    {\(\mathbf{F}_{j}=\nabla E_{\mathrm{DFT}}(\mathbf{x}_{j})\)\\[-1pt]
     \(\mathbf{x}_{j+1}=\mathbf{x}_{j}+\eta\mathbf{F}_{j}\)};

\end{tikzpicture}

\caption{Common two-stage relaxation workflow: an initial structure is first relaxed for \(k_{MLFF}\) steps using an MLFF-driven force update loop. After MLFF convergence, the resulting structure initializes a second relaxation loop of \(k_{DFT}\) steps using DFT-computed forces, producing the final relaxed configuration.} \label{fig:relaxPipeline}
\end{figure*}

First-principles electronic-structure methods provide accurate predictions of energies, and forces, but their substantial computational cost limits the quantity and size of systems that can be practically computed. Machine learning force fields (MLFFs), also referred to as machine learning interatomic potentials (MLIPs), offer an attractive alternative by learning to approximate Density Functional Theory (DFT) from collected datasets. MLFFs can provide predictions several orders of magnitude faster than DFT, enabling high-throughput screening and significantly reducing the cost of structural relaxations. Often, it is used a pre-relaxation step, as seen in Figure \ref{fig:relaxPipeline}.

State-of-the-art architectures and continuing research in MLFFs have been dominated by graph-based and equivariant neural network architectures \cite{bronstein2021geometric, reiser2022graph}, which have demonstrated strong performance across bulk materials and molecular benchmarks \cite{batatia2022mace, deng2023chgnet, yang2024mattersim, frank2024euclidean, poltavsky2021machine, chen2022universal, choudhary2021atomistic, schutt2017schnet, batzner20223, musaelian2023learning}. Graph neural networks (GNNs), being a subclass of geometric machine learning, encode physically derived symmetries into the model directly. This not only ensures that model predictions will satisfy certain known properties, but increases model efficiency on limited datasets \cite{miller2020scaling, batzner20223, hendriks2025similarity, vignac2020building}. Recent approaches in GNNs have included the creation of specialized attention mechanisms \cite{frank2022so3krates, frank2024euclidean} and higher-order message passing \cite{batatia2022mace, chen2022universal}. These models have also begun to be applied to defect prediction tasks, ranging from formation-energy estimation \cite{rahman2024accelerating, xiang2024exploration, fang2025leveraging} to accelerating first-principles atomic relaxations \cite{mosquera2024machine}. 

Despite this progress, applying GNN-based MLFFs to point-defect systems remains challenging. Inherent to defect computations is a repetition of cells to form a supercell \cite{freysoldt2014first,ivady2018first}, in which only the central cell contains the defect. Large supercells are needed to avoid spurious interactions between defect images due to periodic boundary conditions. The resulting structures often contain hundreds to thousands of atoms, dramatically increasing the cost of message passing and exacerbating oversquashing and oversmoothing effects that degrade the propagation of long-range information \cite{li2018deeper}. An approach that modifies the GNN architecture to place extra emphasis on local neighborhoods \cite{fang2025leveraging} was shown to be effective for properties dominated by short-range interactions, such as defect-formation energies. One-hop relaxation strategies have shown promise in 2D materials \cite{yang2025modeling}, yet such amortized approaches require extremely large and diverse datasets \cite{lopez2024neural, amos2025tutorialamortizedoptimization, qiu2022dimes}, which remains impractical for complex 3D defect configurations.

These limitations motivate architectural alternatives for defect supercells that explicitly model non-local interactions. Inspired by the global receptive fields of Transformer models \cite{vaswani2017attention} which have demonstrated broad effectiveness in natural language processing \cite{zhou2024comprehensive}, computer vision \cite{dosovitskiy2020image}, and biomolecular modeling \cite{abramson2024accurate}, we introduce the \emph{Accelerated Deep Atomic Potential Transformer} (ADAPT). ADAPT is a coordinate-based Transformer that computes full attention over all atomic pairs and predicts per-atom forces directly from Cartesian coordinates and elemental features, thereby circumventing the depth-dependent signal degradation inherent to message-passing architectures. 

As access to high-performance computing increases, it is becoming increasingly viable to generate large datasets for supercell defect relaxations. We train ADAPT on a newly released large PBE-level DFT dataset of silicon point defects, including a broad range of complex defect configurations \cite{Xiong2023sa,xiong2024jacs}, as seen in Section \ref{sec:prelim} and Appendix \ref{sec:dataDetails}. ADAPT achieves state-of-the-art accuracy in both energies and forces, outperforming universal MLFFs such as pretrained MACE \cite{batatia2022mace} and MatterSim \cite{yang2024mattersim}, as well as a MACE model retrained on the same dataset. Notably, ADAPT trains at a cost approximately an order of magnitude lower than comparable graph-based architectures. These results indicate that globally receptive, coordinate-based Transformer models provide a promising direction for large supercell structural relaxations of defects. 

Since the initial release of this work, other studies have independently explored Transformer-based MLFFs without explicit equivariance \cite{kreiman2025transformers, elhag2025learning}, as well as a related line of work developed in parallel and released in stages \cite{eissler2026simplegoofftheshelftransformer}. While the earliest version of this latter work predates ADAPT, its most recent results appeared after our initial release and overlap only partially with our approach. In particular, it relies on symmetry-aware data representations rather than architectural equivariance, whereas ADAPT enforces no symmetry beyond those implicit in the training data. Collectively, these results demonstrate that non-equivariant Transformer-based MLFFs can achieve state-of-the-art accuracy on standard molecular benchmarks.

\begin{figure*}[htbp]
    \includegraphics[width=0.9\textwidth]{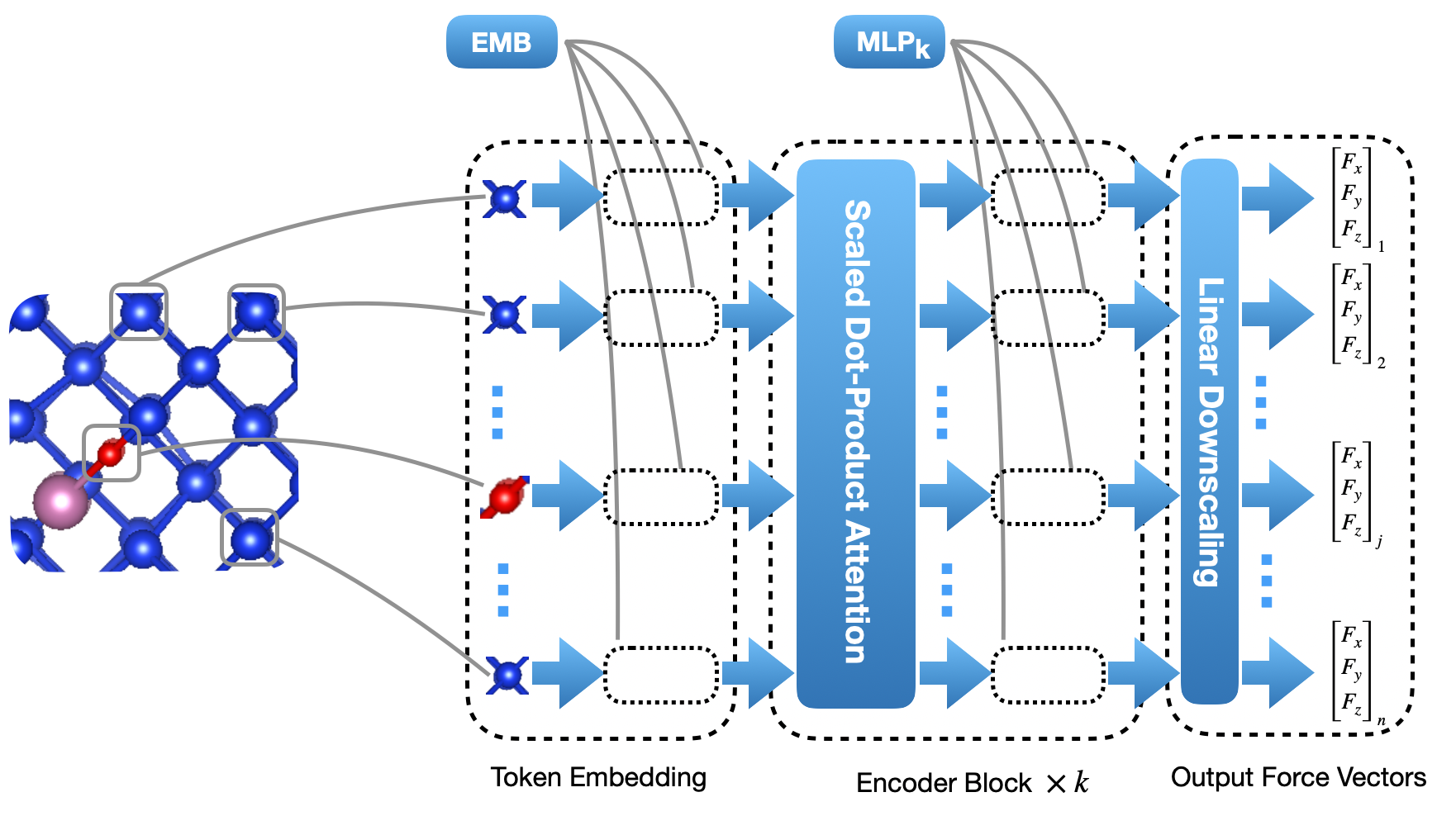} 
    \caption{ADAPT MLFF force-prediction architecture. Following the Transformer encoder paradigm, each of $n$ atoms is embedded via an identical multi-layer perceptron denoted EMB, passed through a stack of Attention-based encoder blocks, and linearly projected from $(n\times d_\text{model})$ to $(n\times3)$ force vectors. }
    \label{fig:MLFFarch}
\end{figure*}

\section{Results} \label{sec:res}
\subsection{MLFFs Without Energy Conservation}
In contrast to MACE \cite{batatia2022mace}, MatterSim \cite{yang2024mattersim}, and related architectures, ADAPT employs separate neural networks for force and energy prediction. While ADAPT's force and energy models each outperform state-of-the-art baselines on the target benchmarks, this design breaks strict energy conservation, since the predicted forces are not constrained to equal the gradient of a scalar potential energy surface. Conservative formulations have historically been viewed as essential for physically faithful atomistic simulation, particularly for long-timescale molecular dynamics (MD) and thermodynamic sampling \cite{chmiela2017machine, batatia2022mace}. However, several recent works have questioned whether strict conservation is always necessary for downstream tasks such as structure relaxation, screening, or surrogate optimization \cite{zhang2024pretraining, bigi2024dark, fu2022forces}. In particular, direct-force or non-conservative models can substantially reduce computational overhead by avoiding backpropagation through atomic coordinates during inference, while also improving force prediction accuracy in some regimes \cite{bigi2024dark, neumann2410orb, rhodes2025orb, eissler2026simplegoofftheshelftransformer}. Recent analyses demonstrate that in some cases, such approaches may still remain usable in MD settings, where instability and energy drift are mitigated by the high accuracy of  the models \cite{eissler2026simplegoofftheshelftransformer}, and others advocate hybrid conservative/non-conservative approaches \cite{bigi2024dark}. Despite these works, ADAPT is not intended as a general-purpose MD force field. Instead, it is explicitly created and verified for rapid structure relaxation and high-throughput defect screening, where the primary objective is accurate convergence to relaxed geometries rather than faithful reproduction of relaxation dynamics. 

The energy-only predictor of ADAPT also serves several independent purposes. First, despite the overall ADAPT framework being non-conservative, the energy model can still be paired with the force predictor during downstream optimization procedures such as FIRE, where it demonstrates superior practical relaxation performance compared to a state-of-the-art conservative baseline, as seen in Section \ref{sec:relaxRes}. Second, standalone energy prediction models are broadly useful for high-throughput materials discovery workflows, including materials stability estimation, candidate filtering, ranking, and rapid surrogate evaluation prior to expensive first-principles calculations \cite{schutt2018schnet, unke2021machine, zhang2024pretraining}. 

\subsection{Architecture Preliminaries} \label{sec:prelim}
Both the proposed force and energy predictor architectures eschew graphs and inductive biases entirely, instead focusing on precise representations of geometries. Our primary aim is to develop force and energy predictors tailored for defect computations, with the longer-term objective of reducing reliance on costly DFT steps in relaxations.

ADAPT force and energy architectures adopt the now standard tokenization paradigm \cite{webster1992tokenization} from Transformer-based deep learning \cite{vaswani2017attention} of breaking inputs into sequences of \emph{tokens}. Here, each token corresponds to a single atom, so a structure with $n$ atoms is represented by $n$ tokens. Initially present in the dataset is a 12-dimensional vector representing each atom by: 
\[
(x,y,z,\text{column},\text{row},\chi,r_{\text{cov}},N_{\text{val}},E_{\text{ion}_1},E_{\text{EA}},r_{\text{atom}},V_{\text{mol}}),
\]
where we define $x, y, z$ as the coordinates of the atom, \textit{column} is the atom's group, \textit{row} is the atom's period, $\chi$ is the electronegativity, $r_{\text{cov}}$ is the covalent radius, $N_{\text{val}}$ is the number of valence electrons, $E_{\text{ion}_1}$ is the first ionization energy of the atom, $E_{\text{EA}}$ is the electron affinity, $r_{\text{atom}}$ is the atomic radius, and $V_{\text{mol}}$ is the molar volume. Experiments in Appendix \ref{sec:descriptSelect} show that there is a negligible difference between using the set of 12 features and the common approach of using just the coordinates and atomic number \cite{batatia2022mace, deng2023chgnet, chen2022universal}. 

ADAPT has been designed to predict the forces and energy for structures that are simulated on computations in a supercell. We consider defect in silicon as our motivating example. DFT calculations were carried out using PBE functional \cite{perdew1996generalized} with a 216 atom-site supercell with PAW pseudopotential \cite{Xiong2023sa,xiong2024jacs}. The dataset is composed of roughly $6{,}000$ defect configurations in neutral charge state. Further details on the dataset creation are available in Appendix \ref{sec:dataDetails}. 

\subsection{Force-Prediction Methodology}\label{sec:method}
Herein, we consider the model architecture used to predict per-atom force vectors, as shown in Figure \ref{fig:MLFFarch}. It can be viewed as a function mapping each token to a corresponding force vector.

\noindent \textbf{Embedding.}\label{par:embed} 
Rather than working in the native 12-dimensional space, we embed each token into a higher-dimensional space of size $d_\text{model}$ (a user-set hyperparameter). High-dimensional representations allow neural networks to transform complex nonlinear dynamics into latent spaces in which linear or low-complexity nonlinear transformations are sufficient to approximate the underlying oracle function, defined as the true generative process that produces the observed real-world data.

A multi-layer perceptron (MLP)~\cite{cybenko1989approximation} is used to learn the embedding transformation, and can be represented as: 
\begin{equation}
\texttt{MLP}(\mathbf{x}) 
= \mathbf{W}_k \sigma \Bigl(\mathbf{W}_{k-1}   \sigma\bigl(\dots\sigma(\mathbf{W}_0 \mathbf{x} + \mathbf{b}_0)\dots\bigr) + \mathbf{b}_{k-1} \Bigr) + \mathbf{b}_k,
\end{equation}
where $\mathbf{x} \in \mathbb{R}^{12}$ is the input token, $\sigma$ is the element-wise ReLU operation defined by: ReLU$(x) = \max(0, x)$, $\mathbf{b}_j \in \mathbb{R}^{d_{\text{out}, j}}$ are the trainable bias terms, and $\mathbf{W}_j \in \mathbb{R}^{d_{\text{out}, j} \times d_{\text{in}, j}}$ are learnable weight matrices. 
Here $d_{\text{in}, 0} = 12$, and $\mathbf{W}_k \in \mathbb{R}^{d_\text{model} \times d_{\text{out}, k-1}}$. 
The embedding MLP is applied independently to each token. 

\subsubsection{Transformer Encoder}\label{sec:encoders}
The embedded sequence is processed by $k$ encoder blocks. Each block has the same structure but distinct parameters. A block is defined by:
\begin{align}
    \mathbf{H}_1 &= \texttt{LN}\bigl(\mathbf{X}_{\text{in}} + \texttt{Attn}(\mathbf{X}_{\text{in}})\bigr), \\
    \mathbf{H}_2 &= \texttt{FFN}\bigl(\mathbf{H}_1\bigr), \\
    \mathbf{X}_{\text{out}} &= \texttt{LN}(\mathbf{H}_2 + \mathbf{H}_1).
\end{align}

The main components are:

\medskip
\noindent \textbf{$(i)$ Layer Normalization ($\texttt{LN}$).} This is used to ensure numeric stability in training, and prevent the chaining together of multiplied terms from growing or shrinking rapidly. 
Given an input $\mathbf{x} \in \mathbb{R}^H$, layer norm normalizes across feature channels:
\begin{align}
    \mu &= \tfrac{1}{H} \sum_{i=1}^H x_i, 
    & \sigma^2 &= \tfrac{1}{H} \sum_{i=1}^H (x_i - \mu)^2, \\
    \hat{x}_i &= \frac{x_i - \mu}{\sqrt{\sigma^2 + \epsilon}}, 
    & y_i &= \gamma_i \hat{x}_i + \beta_i, \quad i = 1,\dots,H,
\end{align}
where $\boldsymbol{\gamma}, \boldsymbol{\beta} \in \mathbb{R}^H$ are learnable parameters and $\epsilon$ is a small constant for stability.

\medskip
\noindent \textbf{$(ii)$ (Multiheaded) Scaled Dot-Produce Attention ($\texttt{Attn}$).} In the model, this is the only place where the tokens interact and influence each other. In multiheaded attention, each ``head'' performs an Attention operation over a subset of the data. Given $\mathbf{X} \in \mathbb{R}^{n \times d_\text{model}}$ (sequence length $n$), each head $i=1,\dots,h$ is defined by:
\begin{equation}
\text{head}_i = 
\operatorname{softmax} \left(\frac{\mathbf{Q}_i \mathbf{K}_i^{\mathsf{T}}}{\sqrt{d_k}}\right)\mathbf{V}_i,
\end{equation}
where
\begin{equation}
\mathbf{Q}_i = \mathbf{X}\mathbf{W}_{\mathbf{Q}_i}, \quad
\mathbf{K}_i = \mathbf{X}\mathbf{W}_{\mathbf{K}_i}, \quad
\mathbf{V}_i = \mathbf{X}\mathbf{W}_{\mathbf{V}_i},
\end{equation}
with projection matrices $\mathbf{W}_{\mathbf{Q}_i}, \mathbf{W}_{\mathbf{K}_i}, \mathbf{W}_{\mathbf{V}_i} \in \mathbb{R}^{d_\text{model} \times d_k}$.  
The raw similarity matrix $\mathbf{Q}_i \mathbf{K}_i^{\mathsf{T}} \in \mathbb{R}^{n \times n}$ encodes pairwise token similarities. The row-wise softmax, where $\text{softmax}(\mathbf{z}_i) = \frac{e^{\mathbf{z}_i}}{\sum_{j=1}^n e^{\mathbf{z}_j}}$, maps each row into a probability distribution over tokens.  

Outputs from all heads are concatenated and projected:
\begin{equation}
\texttt{Attn}(\mathbf{X}) 
= \operatorname{Concat}(\text{head}_1,\dots,\text{head}_h)\mathbf{W}_O,
\end{equation}
with $\mathbf{W}_O \in \mathbb{R}^{h d_k \times d_\text{model}}$.

\medskip
\noindent \textbf{$(iii)$ Feed-Forward Network ($\texttt{FFN}$).} FFNs are applied to individual tokens independently, and do not allow any interactions between tokens. They allow for expressive transformations of the token beyond what Attention alone can capture.
They are given by a position-wise MLP, applied identically to each token:
\begin{equation}
\texttt{FFN}(\mathbf{H}) = \mathbf{W}_2   \operatorname{ReLU} \bigl(\mathbf{W}_1 \mathbf{H}^\mathsf{T} + \mathbf{b}_1\bigr) + \mathbf{b}_2,
\end{equation}
where
\[
\mathbf{H} \in \mathbb{R}^{n \times d}, \quad
\mathbf{W}_1 \in \mathbb{R}^{d_{\text{ff}} \times d}, \quad
\mathbf{W}_2 \in \mathbb{R}^{d \times d_{\text{ff}}},
\]
\[
\mathbf{b}_1 \in \mathbb{R}^{d_{\text{ff}}}, \quad
\mathbf{b}_2 \in \mathbb{R}^{d}.
\]

\medskip
\noindent \textbf{$(iv)$ Dropout.} 
Dropout randomly masks neuron activations (set to $0$), resampled at each pass during training. This has been shown to prevent models from overfitting to the data, and improve generalizability \cite{srivastava2014dropout}. It is applied to the outputs of attention and feed-forward layers. Following convention, we exclude it from the equations for the model definition since it is only used during training and not inference.

\medskip
\noindent \textbf{Force Projection.} 
Finally, after the encoder blocks, forces are obtained by a linear projection:
\begin{equation}
\mathbf{\widehat{y}} = \mathbf{X}_{\text{enc}} \mathbf{W}_{\text{out}}, 
\quad \mathbf{W}_{\text{out}} \in \mathbb{R}^{d_\text{model} \times 3},
\end{equation}
producing per-token force vectors $(f_x, f_y, f_z)$. The resulting tensor has shape $n \times 3$.  

\subsubsection{Defect-Aware Loss Function} \label{sec:lossFxn}

Training requires a differentiable objective that captures the mismatch between predicted and true atomic forces. 
A natural baseline is the mean‑squared error (MSE). Plain MSE, however, does not bias towards any one atom implicitly, even though domain knowledge tells us that atoms nearest the defects dominate the crystal's mechanical response. 

To emphasize these critical regions, we introduce a new loss function: \textit{importance‑weighted MSE.}
In particular, we create an importance mask $\mathbf{m} \in\mathbb{R}_{+}^{n}$, where each of the $n$ atoms, $a_i$, receives weight:
\begin{equation}
  \label{Eq:ImportanceWeightedMSEWeight}
  m_{i} = 
  \prod_{j\in\mathcal{D}}
  \Bigl(1+\frac{\lambda_{1}}
  {\lVert\mathbf{r}_{i}-\mathbf{r}_{j}\rVert^{2}+\lambda_{2}}\Bigr),
  \quad
  \mathcal{{D}}=\{\text{defects}\},
\end{equation}
where $\mathbf{r}_i$ is the coordinate vector for atom $i$, and ${\mathcal{D}}$ is the set of defect locations. The formulation used herein does not consider vacancies, but could easily be modified to do so if necessary. This is similar to laws observed in nature, where the effect of many interactions decay as a power law of the distance between them \footnote{An alternative weighting would be
\begin{equation}
  \label{Eq:ImportanceWeightedMSEAlternativeWeight}
    \sum \ln{\frac{1}{\lVert\mathbf{r}_{i}-\mathbf{r}_{j}\rVert^{2}+\lambda_{2}}}.
\end{equation}
We experimented with Eq.\ (\ref{Eq:ImportanceWeightedMSEAlternativeWeight}), but found that for silicon defects Eq.\ (\ref{Eq:ImportanceWeightedMSEWeight}) gave better results.
It is possible that Eq.\ (\ref{Eq:ImportanceWeightedMSEAlternativeWeight}) would perform better in some applications.}. Hyperparameters $\lambda_1, \lambda_2$ are used to ensure numerical stability and to ``temper'' the scaling. 
The resulting loss becomes:
\[
  \mathcal{L}(\mathbf{\widehat{y}}, \mathbf{y}) = \sum_i m_i \sum_{j} ({\widehat{y}}_{i, j} - {y}_{i, j})^2,
\]
where $\mathbf{y}, \mathbf{\widehat{y}}$ are the actual and predicted forces for each of the atoms (indexed $i$) and across each of the $3$ components of the force vectors (indexed $j$). Comparisons of the effect of the custom loss weighting component against three trials of standard MSE loss show the custom loss weighting version achieves roughly \( 12.5\%\) less error on average than regular MSE. While this loss function is designed specifically for defects, other weighting rules can be designed to match other datasets and training scenarios. 

\subsection{Energy Prediction Methodology} \label{sec:nrg}
We train a separate energy-predictor model to complement the force prediction ADAPT MLFF. For this task, we consider three distinct architectures: (1) a specialized decoder (see Appendix \ref{sec:decoder} for architectural details), (2) a multilayer perceptron (MLP) (see Section \ref{par:embed}), and (3) an MLP$+$residual network. In each case, the model receives only the atomic structure and returns an estimated crystal energy. Architectures (1) and (2) serve as natural baselines; the decoder as a single-output is the natural extension of the encoder framework, and the MLP is a widely used approach \cite{jha2018elemnet, liang2023universal, zhang2018deep}. Architecture (3), however, substantially outperforms both, and we adopt it as our primary design.

\subsubsection{MLP$+$Residual Architecture}
Residuals connections, where the input and output of a layer are added together, have become widespread in ML literature. It has been noted that the residual architecture bears striking resemblance to Euler integration \cite{muller2019space, baggenstos2021approximation} making it a common choice \cite{moghaddam2025advanced, noorizadegan2024stable, kashinath2021physics} when considering modeling physical systems which are governed by differential equations. The architecture of a MLP with residual connections for raw input tokens $\mathbf{x}$ is:

\begin{align*} \label{eq:resid}
    \mathbf{t}_0 &= \sigma(\mathbf{W}_{0} \mathbf{x} + \mathbf{b}_0)\\
    \mathbf{h}_0 &= \texttt{LN}(\mathbf{P}_{0} \mathbf{t}_0 + \mathbf{t}_0)\\
    \mathbf{t}_1 &= \sigma(\mathbf{W}_{1} \mathbf{h}_0 + \mathbf{b}_1)\\
    \mathbf{h}_1 &= \texttt{LN}(\mathbf{P}_{1} \mathbf{t}_1 + \mathbf{t}_1)\\
    &~~\vdots \\
    \mathbf{\widehat{y}} &= \mathbf{W}_k \mathbf{h}_k + \mathbf{b}_k
\end{align*}
\noindent where $\mathbf{W}_i, \mathbf{P}_i, \mathbf{b}_i$ are learnable weight matrices/vectors of any mathematically valid dimensions. However, if $\texttt{Dim}(\mathbf{h}_{i-1}) = \texttt{Dim}(\mathbf{h}_i)$, then $\mathbf{P}=\mathbf{I}$. Dropout (see Section \ref{sec:encoders}) is applied after each ReLU activation function $\sigma$, and all other notation matches that used in Section \ref{sec:encoders}.
Unlike Transformers, MLPs and MLP$+$residuals, require fixed‑length inputs. Based on the structures present in our data, we pad, i.e: add dummy atoms where all values are 0, to every structure to 220 atoms before feeding it to the network. The selection of 220 atoms stems from the regular Si lattice box in the dataset having $6^3 = 216$ atoms, and allowance for the inclusion of point defects. For larger systems, the energy-predictor model can be retrained or fine‑tuned with a higher maximum length rather than truncating atoms.

\begin{table}[t]
\caption{Selection Performance}
\label{tab:energyResults}
\begin{ruledtabular}
\begin{tabular}{lc}
Model Type & Total $\mathcal{L}_2$ Error \\
\hline
Decoder & 23.5508 \\
MLP Only & 50.3728 \\
MLP + residual & 11.1683 \\
\end{tabular}
\end{ruledtabular}
\end{table}

\subsubsection{Initial Model Selection}
To quantify performance, we train each candidate for 200 epochs, and evaluate the total loss on the test set. The results are shown in Table \ref{tab:energyResults}. The MLP$+$residual achieves the lowest error, justifying its selection as the recommended architecture. After adopting it, we further refine the model with an additional 200 epochs of training until convergence was observed.

\begin{figure*}[htbp]
    \centering
    \includegraphics[width=\textwidth]{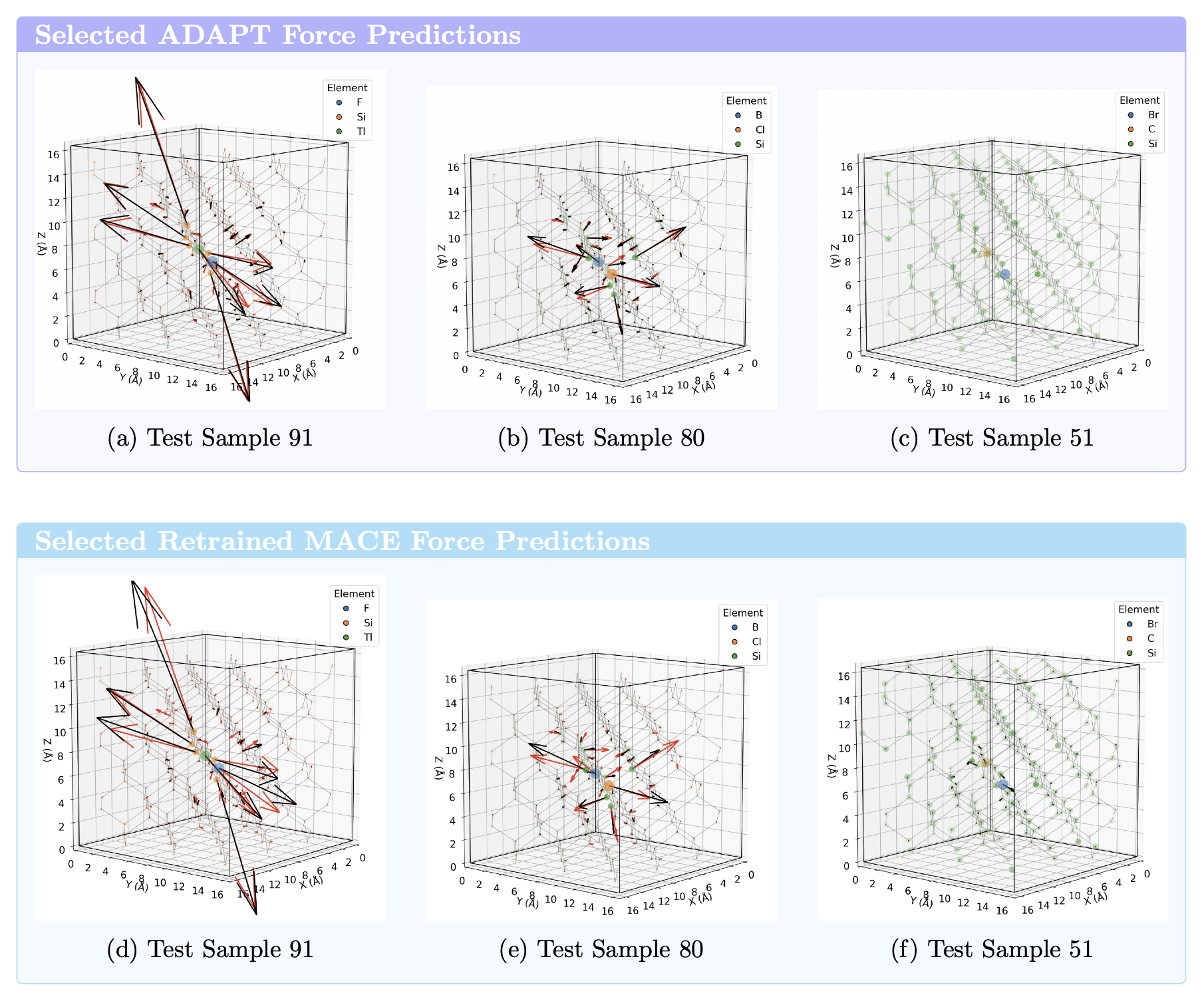} 
    \caption{\centering Side-by-side comparison of instructive outputs. Top row: ADAPT. Bottom row: MACE retrained on the data used to train ADAPT. Predicted forces are shown in black, actual forces are shown in red. }
    \label{fig:model_ab_grid}
\end{figure*}

\subsection{Numerical Results} \label{sec:results}
The primary criterion for comparing MLFFs is accuracy in force and energy prediction, typically measured by $\mathcal{L}_2$ or MAE. 
We benchmark ADAPT against two state-of-the-art models: MACE \cite{batatia2022mace} and MatterSim \cite{yang2024mattersim}. 
To ensure comparability, we train both MACE and ADAPT from scratch on a dataset of $6{,}082$ charge neutral silicon defect DFT trajectories from our previous works, which contains both simple and complex defects with a total of 56 elements \cite{xiong2024jacs,Xiong2023sa}, and select a set of complex defects as the testing scenario. Details of DFT calculations are provided in Appendix \ref{sec:dataDetails}. We additionally report results from previously benchmarked MACE models \cite{batatia2023foundation}. For MatterSim, which is positioned as a large-scale foundation model, retraining is computationally prohibitive; we therefore evaluate using its publicly released checkpoints. All models are evaluated on 100 structures whose trajectories were excluded from training. Performance is assessed in terms of both force prediction accuracy and the ability of ADAPT to recover correct final relaxed structures. For force evaluation, only the initial configurations from each trajectory are considered, as they contain a substantial fraction of non-negligible force components. As structures approach equilibrium, an increasing number of forces decay toward zero, causing $\mathcal{L}_2$-based error metrics to favor trivial or uninformative predictions. This bias arises because most atoms in the bulk lattice undergo negligible displacement, enabling a model to minimize error by globally suppressing atomic motion while failing to capture the small but physically critical displacements that govern structural evolution. Restricting evaluation to initial configurations mitigates this effect by emphasizing larger force magnitudes.
The stability of ADAPT throughout the entire relaxation process is demonstrated by its ability to produce complete and accurate structural relaxations, underscoring its practical utility within atomistic relaxation pipelines.

\begin{figure*}[htbp]
    \centering
    \includegraphics[width=\textwidth]{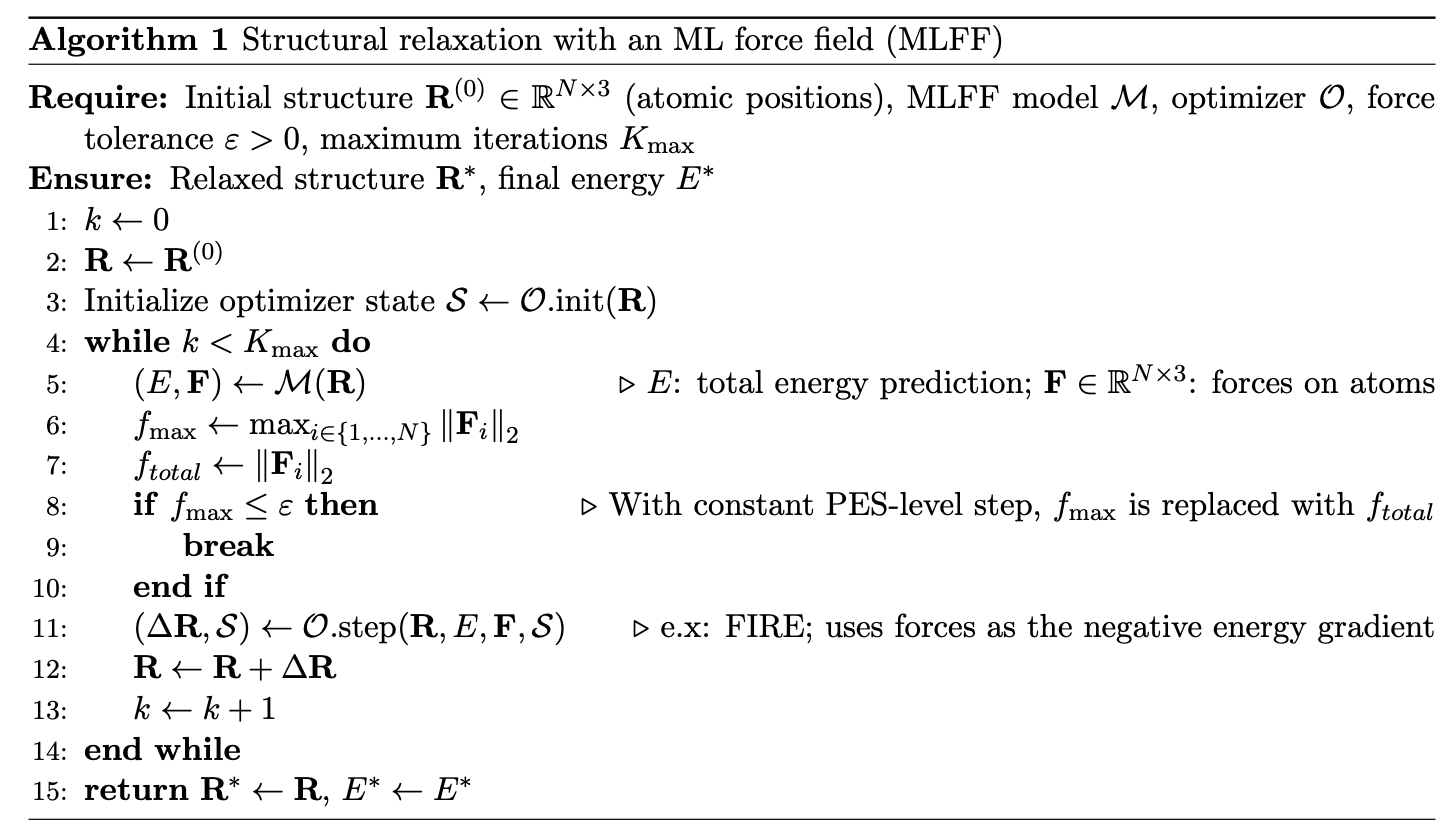} 
    \label{fig:algorithm}
\end{figure*}

\subsubsection{Prediction Accuracy Results}
\begin{figure*}[htbp]
    \centering
    \includegraphics[width=\textwidth]{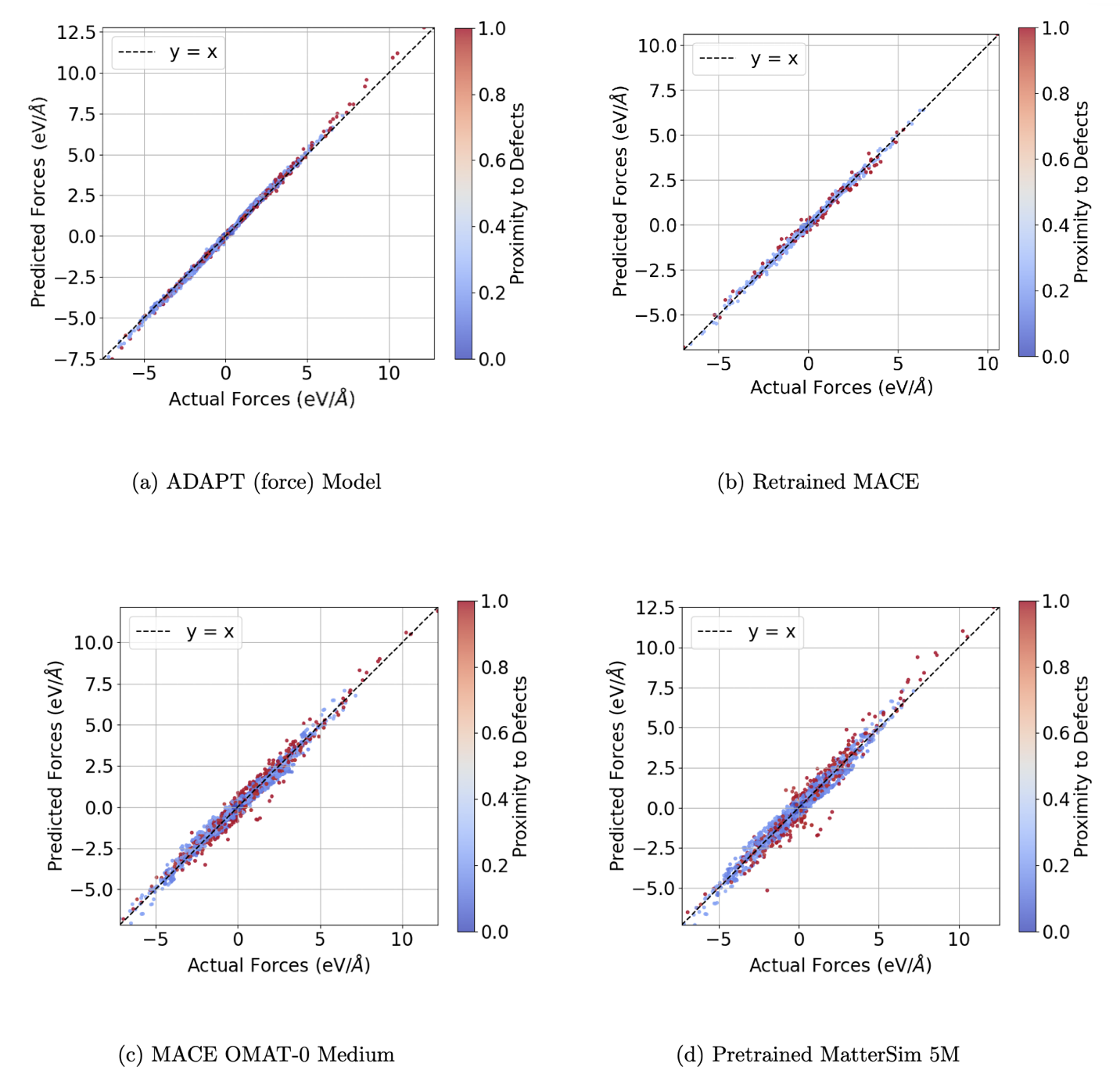} 
    \caption{\centering Parity plots of predicted vs.\ actual forces across test structures. }
    \label{fig:combined}
\end{figure*}

\noindent \textbf{Force Predictions.}
Force prediction results summarized in Table \ref{tab:forceRes} show ADAPT achieves a $25\%$ error reduction relative to retrained MACE, and far outperforms the strongest pretrained model. Parity plots of force and energy errors across the selected test set are shown in Figure \ref{fig:combined}, and examples visualizing the effect on selected structures are included in Figure \ref{fig:model_ab_grid}. The accuracy in forces obtained with ADAPT is around 0.01~eV/\text{\AA} as MAE. This is in the order of magnitude of the stopping criteria for many atomic relaxation within DFT including in our data set. This indicates that ADAPT could be a good surrogate to DFT relaxation and at least provide useful pre-relaxation. 

\begin{table}[t]
\caption{Force prediction comparison with SOTA on 100 Test Structures}
\label{tab:forceRes}
\begin{ruledtabular}
\begin{tabular}{lc}
Architecture &
\begin{tabular}[c]{@{}c@{}}
Force MAE \\
(eV/\AA)
\end{tabular} \\
\hline
\textbf{ADAPT Small} & \textbf{0.0126}\\
MACE Retrained & 0.0161 \\
MACE MP0a Large & 0.0439 \\
MACE MPA-0 Medium & 0.0349 \\
MACE OMAT-0 Medium & 0.0283 \\
MatterSim 1M & 0.0323 \\
MatterSim 5M & 0.0335\\
\end{tabular}
\end{ruledtabular}
\end{table}

\begin{figure*}[htbp]
    \centering
    \includegraphics[width=\textwidth]{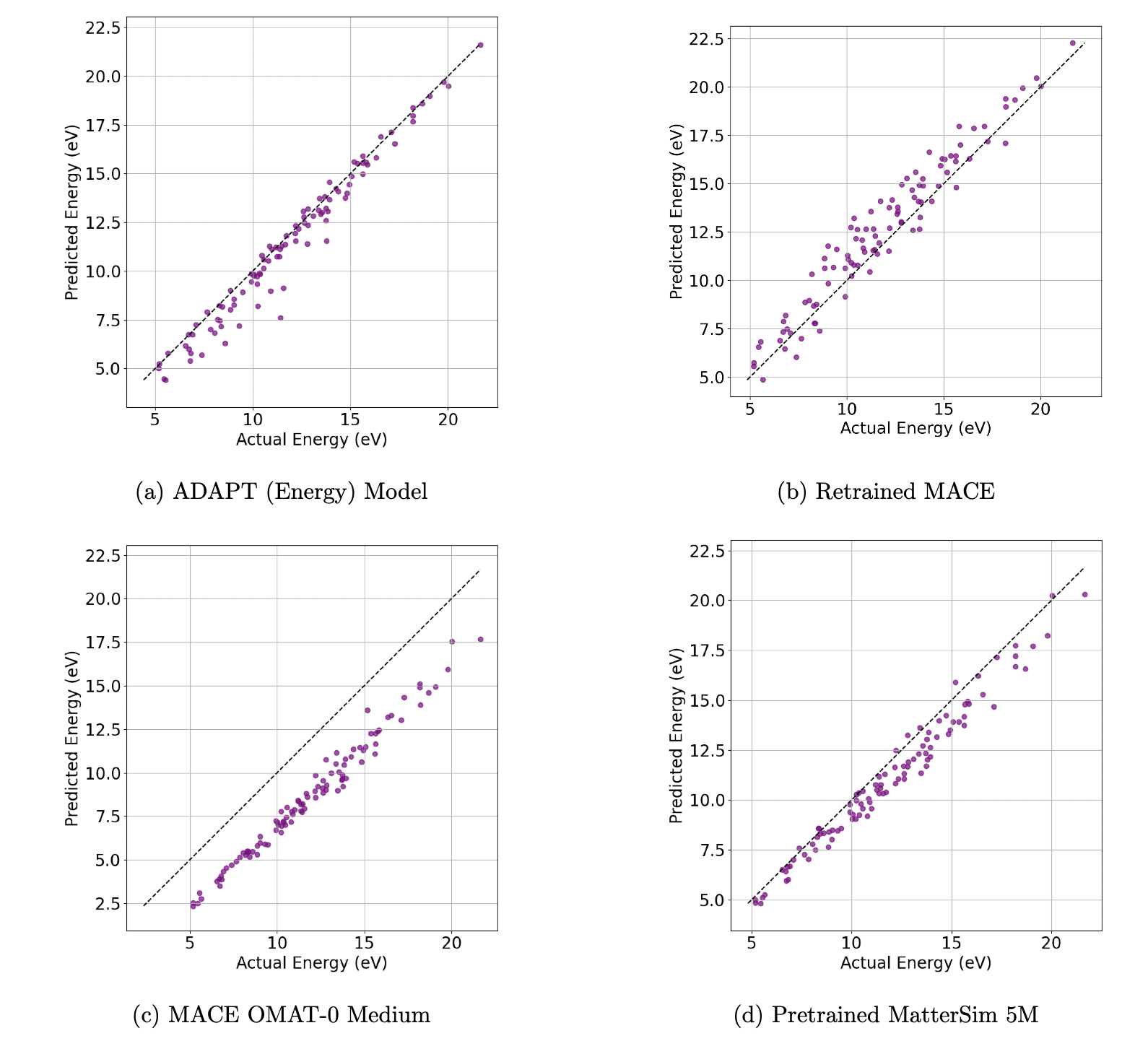} 
    \caption{\centering Parity plots of predicted vs.\ actual defect formation energies across test structures.}
    \label{fig:energyComb}
\end{figure*}

\noindent \textbf{Energy Predictions.}
We also show that the ADAPT defect formation energy-predictor model produces performance superior to both MACE and MatterSim. A table of results is given as Table \ref{tab:energyRes}, and parity plots showing the results are given in Figure \ref{fig:energyComb}. We achieve near identical error to MatterSim 5M, the best of the existing energy predictors, after 200 epochs, and reach our final result, a more than $30\%$ reduction in MAE over MatterSim 5M, after 400 epochs. 

\begin{table}[t]
\caption{Energy prediction comparison with SOTA on 100 Test Structures}
\label{tab:energyRes}
\begin{ruledtabular}
\begin{tabular}{lc}
Architecture &
\begin{tabular}[c]{@{}c@{}}
Energy MAE \\
(eV)
\end{tabular} \\
\hline
\textbf{ADAPT} & \textbf{0.5782} \\
MACE Retrained & 1.0031 \\
MACE MP0a Large & 6.1012 \\
MACE MPA-0 Medium & 2.0478 \\
MACE OMAT-0 Medium & 3.2232 \\
MatterSim 1M & 1.7430 \\
MatterSim 5M & 0.8289 \\
\end{tabular}
\end{ruledtabular}
\end{table}

\subsubsection{Relaxation Results} \label{sec:relaxRes}
Recent studies have demonstrated that accuracy in force prediction alone is insufficient to assess the practical utility of a machine-learning force field \cite{maheshwari2025beyond, fu2022forces}. We show that, when evaluated on the downstream task of structural relaxation, ADAPT's improvements in force and energy predictions translate into more accurate relaxed geometries. Structural relaxations are performed by embedding each model within the iterative relaxation loop, defined in Algorithm \ref{fig:algorithm}, where predicted forces are used by an optimizer to update atomic positions until convergence. ADAPT and MACE are compared using the common FIRE optimizer \cite{bitzek2006structural}, where despite having non-conservative force and energy predictions, ADAPT demonstrates empirical success. 

Table \ref{tab:relaxResults} compares ADAPT with one of the retrained versions MACE \footnote{The relaxation uses a version of MACE which achieved a MAE of $0.0217$, rather than the most recent retraining with error $0.0161$. However, the difference in force accuracy is small compared to the difference in relaxation quality.}, yielding an approximately \(50\%\) reduction in the \(\Delta Q\) score. The \(\Delta Q\) metric is effectively a mean-squared error weighted by atomic mass \cite{alkauskas2014first}. Notably, a further study on relaxations conducted in \cite{dramko2025finetuning}, shows ADAPT consistently achieves \(\Delta Q < 6\) using constant PES-level step sizes, surpassing the performance obtained with the FIRE optimizer. We hypothesize that this improvement arises from the non-conservative nature of employing separate models. Additional details regarding hyperparameter selection and reproducibility are provided in Appendix \ref{append:Details}. 

\begin{table}[t]
\caption{Average $\Delta Q$ score after relaxation with FIRE Optimizer}
\label{tab:relaxResults}
\begin{ruledtabular}
\begin{tabular}{lc}
Model & Avg.\ $\Delta Q$ \\
\hline
MACE  & 12.59 \\
ADAPT & 6.73  \\
\end{tabular}
\end{ruledtabular}
\end{table}

The success of ADAPT over MACE in relaxations raises an interesting consideration regarding energy conservation; namely, that energy conservation alone does not guarantee correct relaxation dynamics. Rather, energy-conserving models guarantee only that the predicted trajectories remain internally consistent with the learned potential energy surface. In practice, inaccuracies in the learned surface can still produce trajectories that are physically consistent, yet ultimately incorrect. In this sense, all MLIPs introduce some degree of physical error, differing primarily in how those errors are distributed. ADAPT makes the explicit tradeoff of reducing overall prediction error at the cost of relaxing the strict coupling between force and energy predictions. Although this sacrifices formal energy conservation, our empirical results suggest that the resulting improvements in predictive accuracy lead to substantially improved structural relaxation performance for the target application considered in this work.

\begin{table}
\caption{\centering Results of data-augmentation training. All rotations are performed about the centroid rather than the origin (or another fixed point). The reported error is the total $\mathcal{L}_2$ error over the test set. Five trials with random rotations are averaged to compute the augmented error. Training approximately invariant models with data augmentation increases computational cost, but can induce approximate equivariance. Appendix \ref{sec:fullAugment} contains detailed studies of performance under rotation and translation augmentation.}
\label{tab:augRes}
\begin{ruledtabular}
\begin{tabular}{cccc}
Aug. Train. & Epochs & Unaugmented Error & Aug. Error \\
\hline
No  & 80  & 4.32 & 121.67 \\
Yes & 80  & 8.92          & 9.00   \\
Yes & 160 & 6.90          & 6.74   \\
Yes & 320 & \textbf{4.17} & \textbf{3.49} \\
\end{tabular}
\end{ruledtabular}
\end{table}

\subsection{Computational Efficiency}
\noindent \textbf{Force Predictions}
An advantage of the ADAPT architecture is its computational efficiency. Table \ref{tab:runtime} shows the per-epoch runtime for training ADAPT and MACE. Both are trained on the $\sim 216$ atom dataset of Si defects. For timing comparison, ADAPT is also trained using simulated noise for $512$, $1024$, and $4096$ ``atom'' dataset sizes. ADAPT computations only depend on the tensor size of the input, not the structure represented, meaning that the processing time for random noise is the same for physical structures. This testing procedure does not carry over to MACE however, where random noise can result in unrealistic numbers of neighbors falling within a cutoff radius. 

Training ADAPT converged after 80 epochs (totaling 3 compute hours). In comparison, training MACE took 60 epochs, amounting to 136 compute hours: more than 45$\times$ the amount of compute used to train ADAPT's force-prediction model on the same dataset. 
The compact design of ADAPT permits training on commodity hardware, including workstations and even consumer-grade laptops equipped with GPUs \footnote{The authors successfully trained ADAPT with $216$ atom supercells on a 2023 M2 MacBook Air laptop.}, thereby significantly reducing hardware requirements for adoption. This accessibility is consistent with the overarching objective of the MLFF literature: to accelerate structural determination by reducing dependence on large-scale computational resources.

\begin{table*}[t]
\caption{Per-epoch runtime. Models are trained on NVIDIA A100 GPUs; average minutes per epoch are reported. Note that while increasing the number of GPUs reduces wall-clock training time, parallel efficiency decreases due to communication, synchronization, and load-imbalance overheads inherent to distributed training.}
\label{tab:runtime}
\begin{ruledtabular}
\begin{tabular}{lccccr}
Model & Atoms & A100 GPUs & Min./Epoch & Batch Size & Compute Min./Epoch \\
\hline
MACE  & 216  & 16 & 8.50  & 8  & 136.00 \\
MACE  & 216  & 1 & 89.40  & 8  & 89.40 \\
\hline
ADAPT & 216  & 1  & 2.24  & 64 & 2.24 \\
ADAPT & 512  & 1  & 7.99  & 64 & 7.99 \\
ADAPT & 1024 & 1  & 25.11 & 48 & 25.11 \\
ADAPT & 4096 & 1  & 341.23 & 2 & 341.23 \\
\end{tabular}
\end{ruledtabular}
\end{table*}

These improvements are attributed to the departure from graph-based architectures. 
Graph neural networks inherently involve sparse operations, which are not easily expressed in the dense linear algebraic form favored by modern accelerators. 
Consequently, graph-based models typically exhibit lower hardware utilization due to sparse operations, which lack the extensive optimization and backend support available with dense-matrix operations \cite{liang2020engn}. 
By forgoing graph representations and adopting architectural paradigms widely developed in natural-language processing and computer vision—where such operations benefit from extensive backend and library support—ADAPT achieves markedly higher computational throughput.

\noindent \textbf{Energy Prediction.}
MACE generates energy predictions concurrently with force predictions within the same forward pass, yielding identical timing characteristics for both quantities. 
ADAPT trains an additional energy-predictor model, which required 1.93 compute hours on a single NVIDIA A100 GPU. 
Model training was conducted for 400 epochs, with the duration of a single epoch being 29 seconds on the same hardware. When including this cost, training both ADAPT models takes a total of 4.92 A100 hours, which is still more than 138$\times$ faster than MACE.

\section{Symmetry Handling Without Graphs}
\subsection{Effect of Rigid Motion Transformations}
\noindent \textbf{Data Augmentation.} Graph neural networks inherently encode three key symmetries of atomistic systems: (i) permutation invariance with respect to atom ordering, and (ii-iii) equivariance under global rigid translations and rotations \cite{bronstein2021geometric, reiser2022graph, deng2023chgnet, yang2024mattersim, frank2024euclidean, poltavsky2021machine, chen2022universal, choudhary2021atomistic, schutt2017schnet, batzner20223, musaelian2023learning}. The encoder in ADAPT has permutation invariance, but it does not impose rotational or translational equivariance. A common approach to ensuring that symmetries like rotational and translational equivariance are respected is to train with \textit{data augmentation}: in which random rotations and translations of the data are used during training. As shown in Table \ref{tab:augRes} and Appendix \ref{sec:fullAugment}, ADAPT can be trained to exhibit a high degree of approximate equivariance to rigid-body transforms: translating or rotating the input structure leads to little change in the accuracy of predicted energies and forces. This mirrors a very general result in machine learning literature referred to as ``the bitter lesson'': with sufficiently large datasets it is very often more beneficial to ``get out of the way'' of a machine learning architecture and let it learn itself than to encode restraints directly into the model \cite{sutton2019bitter}.

\subsection{Scaling Laws and Relaxation Pipeline Efficiency}
ADAPT employs full self-attention, which incurs a computational complexity of $\mathcal{O}(n^{2})$ in the number of atoms $n$. This scaling is inherent to the formulation: explicitly modeling long-range, all-to-all interactions requires considering all atomic pairs, and cannot be reduced without changing the underlying architecture. In this sense, ADAPT intentionally trades asymptotic complexity for expressive capacity.

In contrast, message-passing GNNs typically exhibit $\mathcal{O}(n)$ arithmetic complexity per layer by restricting interactions to a fixed local neighborhood. However, this theoretical advantage does not directly translate to practical performance on modern hardware. GNNs rely on sequential message-passing steps and sparse, irregular memory access patterns, which limit parallelism and result in poor GPU utilization \cite{deng2024mega, auten2020hardware}. Attention layers, by comparison, map naturally to dense linear algebra operations and benefit from highly optimized GPU kernels. As a result, despite its worse asymptotic scaling, ADAPT is often faster in practice. As shown in Table \ref{tab:runtime}, ADAPT has a computational burden an order of magnitude less than MACE on $\sim 217$ atom systems. Even when running systems of $512$ or $1{,}024$ atoms (simulated with the process described in Appendix \ref{append:Details}), ADAPT uses less compute than MACE does for $\sim 217$ atoms systems. 

\noindent \textbf{Benefit compared to DFT.} Even in very large structure regimes where the quadratic scaling of attention leads to longer ADAPT vs GNN times, the tradeoff may remain favorable. A core premise of MLFF-accelerated workflows is that force-field evaluations are negligible compared to ab initio calculations. Standard plane-wave-based Kohn-Sham DFT methods scale as $\mathcal{O}(n^3)$ with system size, and for systems with only $\mathcal{O}(10^{2})$ atoms, MLFFs are already orders of magnitude faster than DFT. Consequently, in a situation in which first-principles refinement is utilized following the MLFF relaxation, any non-trivial improvement in MLFF accuracy that reduces the number of required DFT steps yields a net reduction in total computational cost. In this setting, a slightly more expensive MLFF that converges in fewer DFT iterations is strictly preferable to a faster but less accurate model.

\clearpage

\section{Discussion}
\subsection{Effect of Separate Models on Efficiency}
ADAPT employs separate models for force and energy prediction, a design choice that offers several practical advantages. When only one quantity is required, the corresponding predictor can be deployed independently, reducing both runtime and memory overhead. This distinction is particularly relevant for defect-focused MLFF applications, where simulations are often performed on large supercells containing hundreds of atoms. Such efficiency gains may be important for practitioners operating on local workstations or resource-constrained computing clusters. The separation also improves modularity: force and energy predictors can be retrained or updated independently, enabling the incorporation of datasets that contain only one target quantity and allowing incremental refinements without retraining the full system.


Modularity also introduces optimization-related limitations. Certain relaxation methods, such as the FIRE optimizer \cite{bitzek2006structural}, require simultaneous access to both forces and energies. In these settings, we observe degraded relaxation performance relative to optimizers that depend only on force predictions, as discussed in Section \ref{sec:relaxRes}. Additionally, joint force-energy models are often more parameter-efficient \cite{zhang2021survey}, since they learn a shared representation across tasks and can exploit correlations between forces and energies to improve generalization when sufficient training data are available. We note, however, that conclusions regarding internal neural-network representations should be interpreted cautiously, as the black-box nature of these architectures makes their learned dynamics difficult to characterize directly.


Given the limited training data available for silicon defects, it is not surprising \cite{liu2021efficient, zhu2023understanding, zhang2020you} that a simpler MLP with residual connections outperformed a Transformer decoder in this setting: see Table \ref{tab:energyResults}. Nonetheless, the authors expect that, with sufficient force and energy data, Transformer architectures augmented with specialized heads may provide a more scalable and accurate solution. The design of such heads remains an active area of research, and identifying architectures that best balance modularity, efficiency, and accuracy is an open problem.

\subsection{Coordinates vs. Graphs} 
GNNs are the default backbone for modern MLFFs \cite{batatia2022mace, deng2023chgnet, frank2024euclidean, yang2024mattersim, chen2022universal} where atoms define nodes, and atomic bonds or proximity determine edge placement. 
By encoding geometric priors (permutation, rotation, and translation invariance), they incorporate strong inductive biases that improve data efficiency \cite{bronstein2021geometric, ko2025data, kiechle2024graph, oliva2022graphneuralnetworksrelational} and have been argued to stabilize relaxation trajectories \cite{frank2022so3krates}. 

Representing continuous atomic interactions using discrete graph topologies introduces mismatches that can limit accuracy, especially in defects where long-range effects and precise geometries are important.
GNNs inherently restrict interactions to local regions, relying on network depth to propagate forward information that is outside the interaction radius. 
This approach often leads to over-smoothing and over-squashing \cite{giraldo2023trade, rusch2023surveyoversmoothinggraphneural}, where long-range signals degrade rapidly as depth increases. 
Bulk crystal far from the defect core can substantially shape local defect structures. 
While long-range influences are less critical in many other chemical systems, neglecting them in crystalline materials can cause large errors. 
The poor performance of GNNs on large periodic systems---an issue especially relevant in modeling crystalline defects---has been noted \cite{frank2022so3krates, yan2024case}. 
Adding long-range interactions into graph architectures \cite{frank2022so3krates, frank2024euclidean} often leads to significant cost in computation and model complexity. 
Thus, we arrive at the motivation for using an alternative MLFF strategy for modeling crystal defects in ADAPT, and a need recognized in \cite{frank2022so3krates, yan2024case} as well.

With the advent of Transformer architectures and growing datasets, it is now feasible to move away from hard-coded geometric priors and instead focus on explicit representations of global distances and angles. 
ADAPT employs a Transformer encoder (Section \ref{sec:method}, Appendix \ref{append:Details}) with full, unmasked self-attention, enabling \emph{all-to-all} comparisons between atoms at each layer. 
This approach directly captures non-bonded and long-range interactions without depending on depth-based message passing. 
Although the model lacks explicit geometric equivariances, permutation invariance is inherent to unmasked attention, and experiments show that translational and rotational invariances can be learned sufficiently well from data.
The importance of global attention is underscored in Table \ref{tab:fullAttnImport}: restricting attention to local neighborhoods—as in GNNs—drastically degrades performance.

\begin{table}[t]
\caption{Full versus local interaction. The allowed interaction percentage denotes the fraction of all possible atom--atom interactions retained during training and inference. Interactions are controlled in the attention mechanism via key--structural masks (Appendix~\ref{sec:keyMask}). Lower values indicate lower total $\mathcal{L}_2$ error. Data points marked with ${}^{*}$ converged after 80 epochs, while those marked with ${}^{\dagger}$ were trained for 200 epochs until convergence.}
\label{tab:fullAttnImport}
\begin{ruledtabular}
\begin{tabular}{cc}
Allowed Interactions (\%) & Total $\mathcal{L}_2$ Loss \\
\hline
1.46  & $13.16^{*}$ \\
18.7  & $13.61^{\dagger}$ \\
51.3  & $11.13^{\dagger}$ \\
100   & $8.11^{*}$ \\
\end{tabular}
\end{ruledtabular}
\end{table}

\noindent \textbf{Accurate Representation of Geometries.}
Graphs excel at representing connectivity, but they do not inherently encode precise distances or angular relationships. To address this limitation, many GNN variants augment node and edge features with geometric information \cite{bronstein2021geometric, frank2022so3krates, batatia2022mace, chen2022universal, deng2023chgnet}. However, for long-range interactions this information is only available indirectly, propagated iteratively from neighbor to neighbor. As path lengths between atoms increase, this process can cause signal degradation due to oversmoothing and oversquashing, with errors compounding over distance.

By contrast, a coordinate-based approach gives direct access to precise pairwise distances and angles for all atoms in a single computation step. 
This approach not only avoids approximations from multi-hop propagation, but also preserves geometric detail across all interaction scales. 

\noindent \textbf{Limitations and Future Directions.}
The ADAPT architecture is not inherently limited to defect relaxation or force prediction. However, it remains an open problem to determine ADAPT's applicability to other problems including diverse bulk structures. Additionally, Transformers typically require substantial quantities of data \cite{liu2021efficient, zhu2023understanding, zhang2020you}, making ADAPT unsuitable for tasks with limited training data. The key takeaway of the work however, is that GNN-free MLFFs can reach high accuracy.

Future directions include $i)$ enforcing physical symmetries within both the architecture and the loss while remaining distinct from graph-based techniques; $ii)$ extending training beyond silicon to encompass a wider class of defects and materials; $iii)$ extending the framework to simulate charged defects in semiconductors.

\begin{acknowledgments}
This study was funded by NSF grants CCF-2212558, CCF-2212557, and CCF 1918651. The first principles work has been supported by the U.S. Department of Energy, Office of Science, Basic Energy Sciences in Quantum Information Science under Award Number DE-SC0022289. This research used resources of the National Energy Research Scientific Computing Center, a DOE Office of Science User Facility supported by the Office of Science of the U.S.\ Department of Energy under Contract No.\ DE-AC02-05CH11231 using NERSC award BES-ERCAP0020966. 
The funder played no role in study design, data collection, analysis and interpretation of data, or the writing of this manuscript.
Any opinions, findings, and conclusions or recommendations expressed in this publication are those of the authors, and do not necessarily reflect the views of the sponsoring entities.

This research was funded in part by: The Robert A. Welch Foundation (grant No. C-2118 A.K.); Rice University (Faculty Initiative award); NSF CAREER (award no. 2145629); an Amazon Research Award; a Microsoft Research Award.
\end{acknowledgments}

\section*{Author Contributions}
E.D. created the software, designed methodology, curated data.
Y.Z. trained the MACE models, curated/sourced data, and provided domain expertise. 
Y.X. curated/sourced data, and provided domain expertise. 
C.J. designed methodology
T.R. designed methodology
G.H. provided domain expertise
T.K. designed methodology

All authors contributed to the writing and revisions of the paper. 

\newpage
\appendix
\section{Dataset Details and Code Availability} \label{sec:dataDetails}
\begin{figure*}[t]
    \centering

    \includegraphics[width=\textwidth]{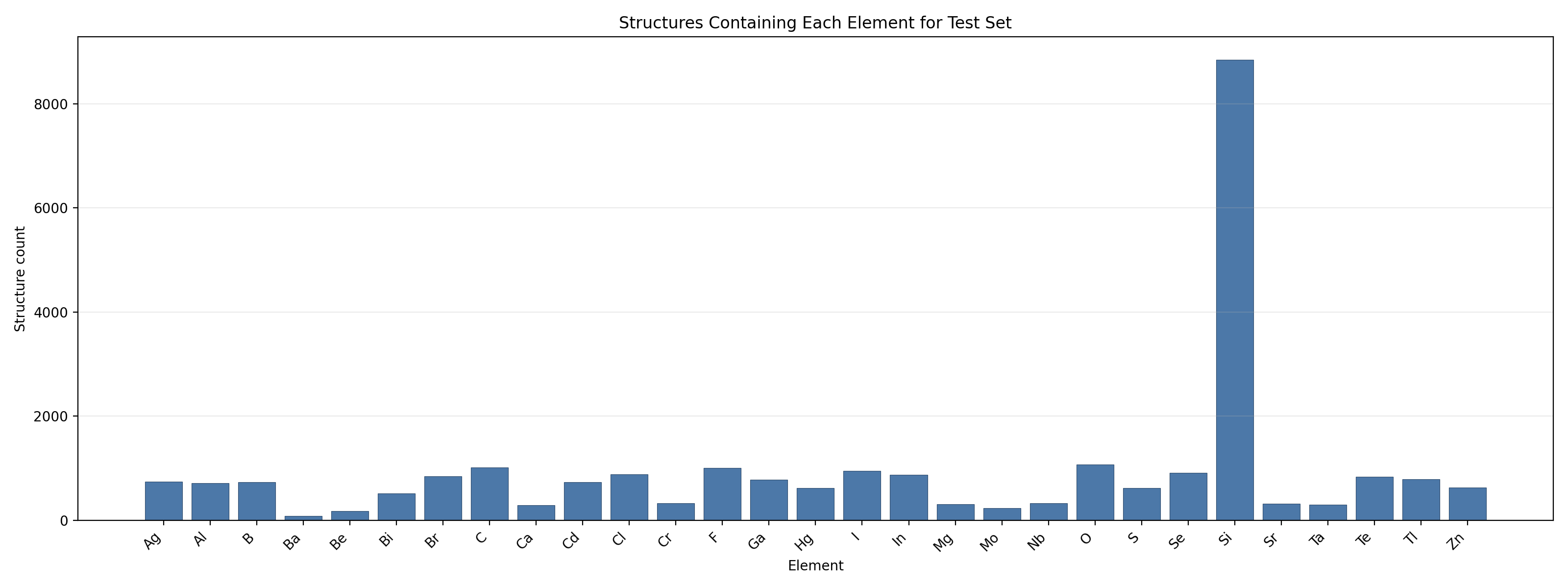}

    \vspace{0.5em}

    \includegraphics[width=\textwidth]{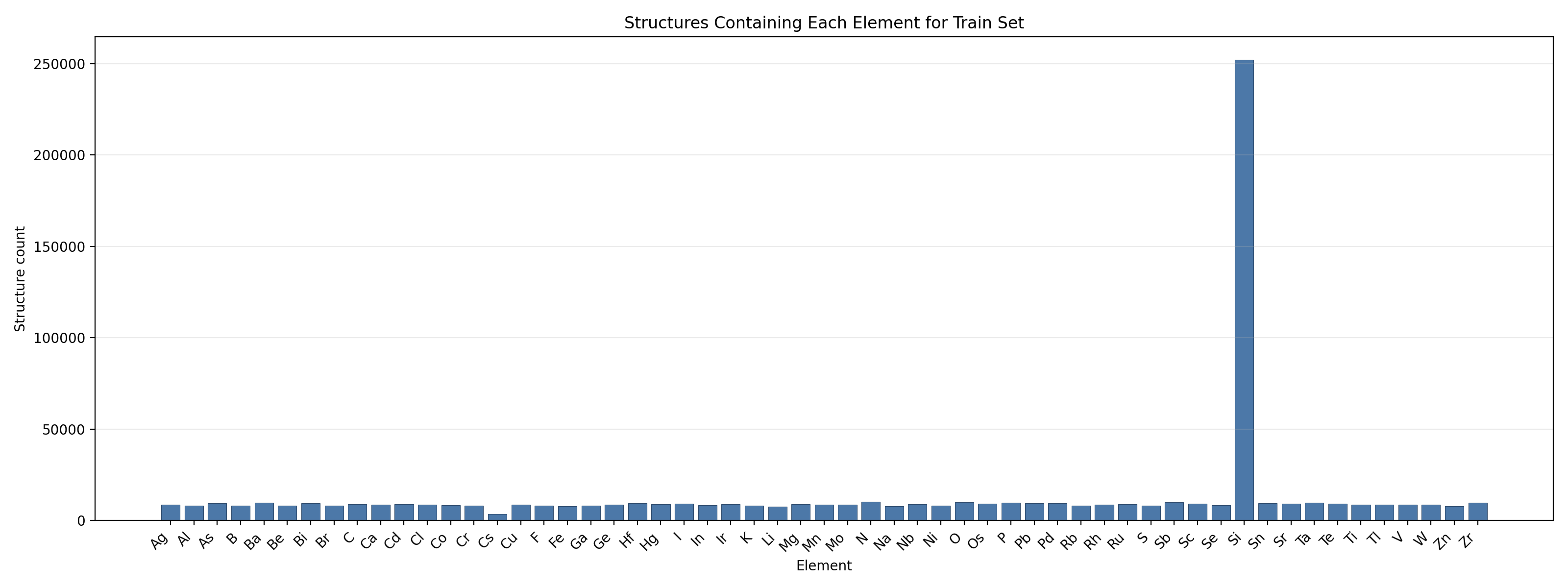}

    \caption{Elemental frequency rate in training and testing datasets.}
    \label{fig:distribCounts}
\end{figure*}

The DFT trajectories dataset contains both simple and complex defects in silicon, which correspond to previous works \cite{xiong2024jacs, Xiong2023sa}. The complex defects are in substitutional-interstitial configuration. The defect elements in the dataset span most of the periodic table besides the noble gas, rare-earth, and the ones that are difficult to implant, giving in total 56 elements \cite{Xiong2023sa}. Charts showing the frequency of the elements in the dataset are included in Figure \ref{fig:distribCounts}. In this work, we extract $252{,}240$ number of single-point calculations of neutral charge defects from the relaxation trajectories. The high-throughput defect computations were performed using the automatic workflows that are implemented in atomate software package \cite{Jain2013, Mathew2017,Ong2013}. The first-principles calculations were performed using Vienna Ab-initio Simulation Package (VASP) \cite{G.Kresse-PRB96,G.Kresse-CMS96} with the projector augmented wave (PAW) method \cite{P.E.Blochl-PRB94}. All the calculations were spin-polarized at the Perdew-Burke-Erzhenhoff (PBE) level\cite{perdew1996generalized}. Defect atoms were embedded in a Si supercell with 216 atoms. 520 eV cutoff energies were used for the plane-wave basis and the Brillouin zone was sampled with single $\Gamma$. All the defect structures were optimized at a fixed volume until the ionic forces were smaller than 0.01~eV/\r{A}.

The datasets generated and/or analyzed during the current study are available in the ``ADAPT\_Released'' Zenodo project with DOI: ``https://zenodo.org/records/17411327''. 

The underlying code and training/validation datasets for this study are available in the GitHub repository: ``ADAPT\_Released'' and can be accessed via the link:
\url{https://github.com/EvanDramko/ADAPT_Released}.

\section{Architecture Details and Hyperparameters} \label{append:Details}
\subsection{Hyperparameters}
\begin{itemize}
    \item ADAPT: We define the ``small'' model size by: [$d_{\text{model}} = 256$, $d_{\text{ff}} = 512$, \#-layers$ = 8$, \#-heads$ = 8$, dropout rate $=0.05$] trained for 80 epochs. The ``large'' model size is: [$d_{\text{model}} = 512$, $d_{\text{ff}} = 1024$, \#-layers$ = 8$, \#-heads$ = 8$, dropout rate $=0.05$] trained for 750 epochs. All training was in single precision. The validation set was the first 500 frames, which helps prevent against data-leakage through series interpolation \cite{bergmeir2018note}. The hyperparameters and training of ADAPT are not necessarily optimal, but are sufficient to demonstrate the key point of the paper that ADAPT-style models can perform competitively with GNNs. 
    \item MACE: The retrained version of MACE (v0.3.14, PyTorch 2.6.0) uses: num\_interactions=2, num\_channels=256, max\_L=2, correlation=3, r\_max=5.0, trained for 300 epochs on single precision (float32). Using early stopping as well, the weights used in this work result from 60 epochs of training, which is where the loss curve begins to exhibit long-tailed properties. 
    \item FIRE Optimizer: All parameters are set to the default parameters of FIRE, with $f_{\max} = 0.05$, and $K_{\max} = 100{,}000{,}000$. 
    \item Constant Step Size Structural Optimizer: details on fitting procedures, hyperparameter selections, and practicality of results are available in \cite{dramko2025finetuning}. Notably, a wide selection of constant step sizes outperform the FIRE based relaxation. 
\end{itemize}

\subsection{Loss Curves and Training}
\begin{figure}[htbp]
    \includegraphics[width=0.48\textwidth]{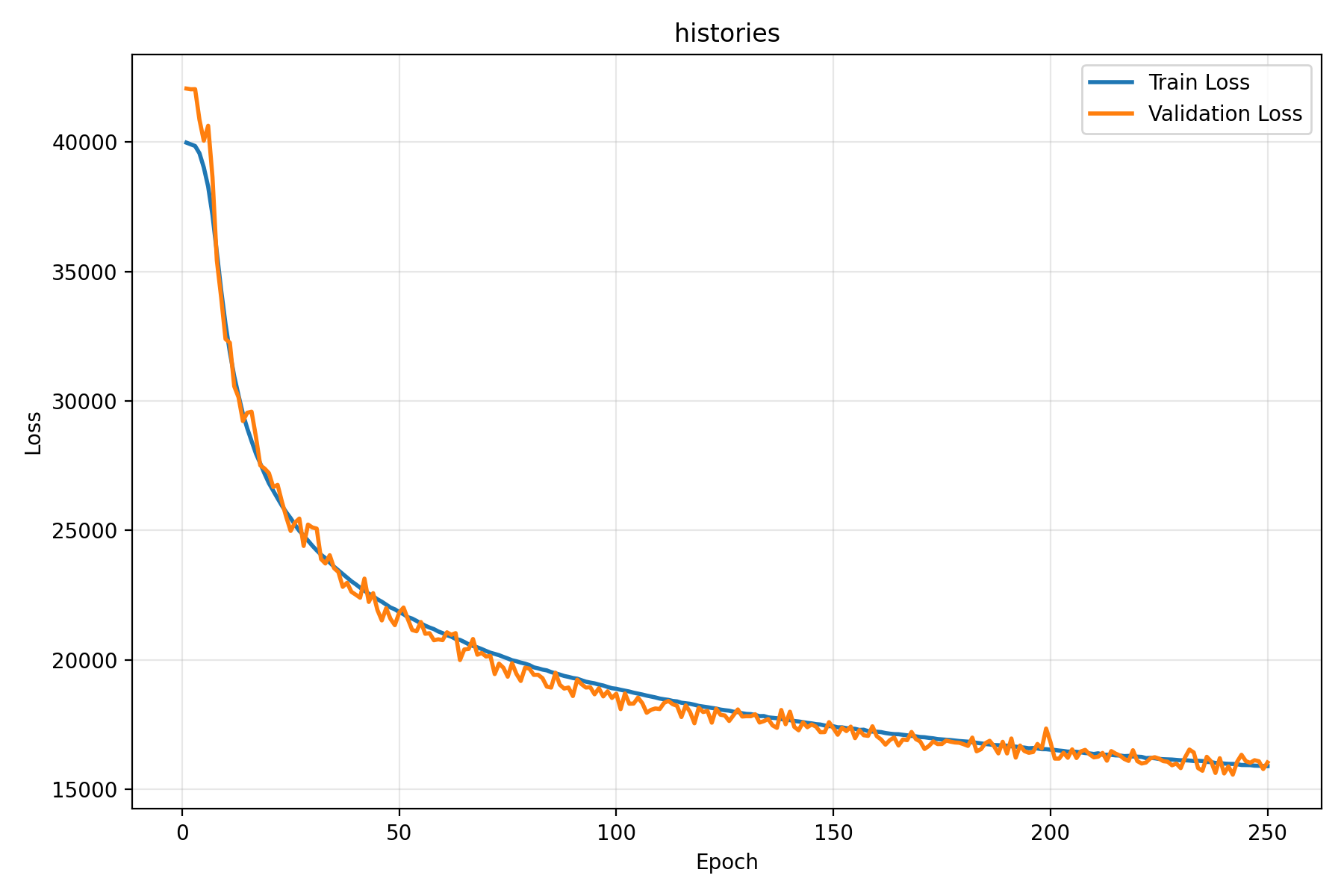} 
    \caption{\centering Loss curves from a sample trial of training the small ADAPT size architecture.}
    \label{fig:adaptCurve}
\end{figure}

\begin{figure}[htbp]
    \includegraphics[width=0.48\textwidth]{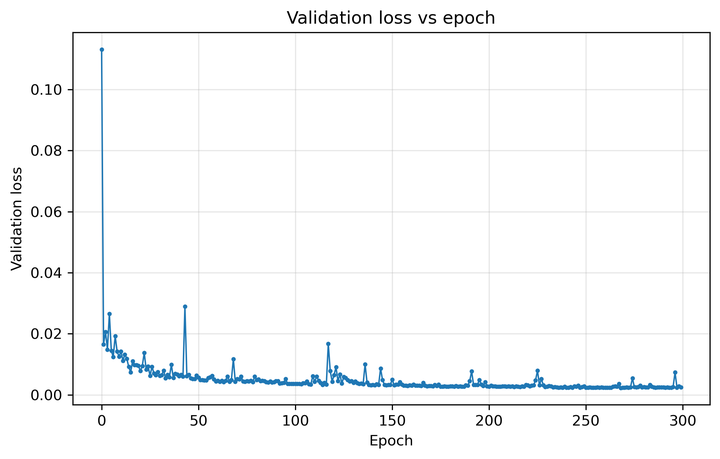} 
    \caption{\centering Loss curve from a sample trial of training the MACE architecture (library only reports validation loss). }
    \label{fig:maceCurve}
\end{figure}

The small ADAPT configuration achieves a MAE of \(0.0120\), whereas the large version resulted in an error of \(0.0136\). Thus, we see that the large configuration exhibited overfitting, indicating that the smaller model already distilled nearly all available information from the data. Accordingly, under the assumption of no additional data or inductive biases, we do not expect training a larger model on the same inputs is likely to achieve a meaningful performance gain. Sample loss curves from ADAPT (small) and MACE training are included in Figure \ref{fig:adaptCurve} and Figure \ref{fig:maceCurve} respectively. 

During ADAPT force prediction training, early stopping 
\cite{earlyStopPrechelt,li2019gradientdescentearlystopping} 
was employed due to the high structural similarity among samples 
and the slow, long-tailed convergence behavior observed in the 
validation loss. The early-stopping criterion was determined 
solely from training and validation performance, without reference 
to the test set. The test set was used only for final evaluation 
after training and hyperparameter selection were completed. 
Post-hoc evaluation on the held-out test set supports this choice: 
the full 250-epoch version of ADAPT-small achieves an MAE of 
\(0.0131\)~eV/\AA{} on the test set, which is 
\(0.005\)~eV/\AA{} worse than the early-stopped model. 
Furthermore, the degraded performance of ADAPT-large relative to 
ADAPT-small suggests potential overfitting on the dataset.

\subsection{Data Simulation}
To simulate large structures of $512$, $1{,}024$, and $16{,}384$ atoms, we create a dummy dataset of $250{,}000$ structures of $(512, 12)$ to represent the structural input, and $(512,3)$ to represent the force output. Because there is not cutoff radius in ADAPT, we are able to use random numbers to fill the tensors without introducing issues. 

\subsection{Evaluation At Different Levels} \label{sec:evalLevel}
While $\mathcal{L}_2$ error is the conventional standard for comparing force predictions, we find that it is insufficient to fully capture the dynamics of point defects in crystals.
To perform a more appropriate comparison, we use two complementary levels. (i) \emph{Model level} (MLFF): accuracy of force and energy predictions. (ii) \emph{Relaxation level}: quality of the final relaxed structure obtained by running a geometry optimizer with the MLFF.

\textbf{Model-Level Evaluation of Forces:} When comparing candidate models, in addition to the loss scores (see Section \ref{sec:lossFxn}), we also consider the average angle and magnitude errors separately. We use the dot product to calculate the angular error in degrees via
\begin{align*}
    \mathit{angle}(\mathbf{y}, \mathbf{\widehat{y}}) &= \arccos\left(\frac{\mathbf{y} \cdot \mathbf{\widehat{y}}}{||\mathbf{y}||_2 \cdot  ||\mathbf{\widehat{y}}||_2}\right) \cdot \frac{180}{\pi},
\end{align*}
and we calculate the difference in magnitudes via
\begin{align*}
    \mathit{mag}(\mathbf{y}, \mathbf{\widehat{y}}) &= \left\lvert \left\lVert \mathbf{y} \right\rVert_2 - \left\lVert \mathbf{\widehat{y}} \right\rVert_2 \right\rvert 
\end{align*}

These results help to determine whether the model is genuinely learning the underlying dynamics or artificially minimizing error by predicting uniformly negligible forces---knowing that in reality, most of them will be close to zero.  Although this angular-magnitude metric is differentiable and theoretically usable as a loss function for the MLFF, in practice it is difficult to balance the angular and magnitude components effectively. Empirical results show that angular-loss functions are often brittle and require significant engineering effort to implement reliably \cite{deng2019arcface}---a result borne out in our own experiments. In contrast, using a weighted mean-squared-error (MSE) loss is simpler, more robust, and yields strong performance at both the MLFF and structural-relaxation levels, making it the preferred choice. 

\textbf{Evaluation of Energy:} The total energy of the crystal is represented with a single number, making evaluation very easy. We use the common $\mathcal{L}_2$ distance metric. 

\textbf{Evaluation of Structural Relaxation} In order to evaluate the final result of the full relaxation procedure, we use the well known SOAP and delta Q metrics. Other checkers (such as those which check bond lengths) are also viable, although we do not use them in this work.

\section{Masking in Attention} \label{sec:keyMask}
When restricting interactions in Attention, we apply \emph{masks} to the attention logit matrix
\[
QK^\mathsf{T} \in \mathbb{R}^{B \times H \times T \times T},
\]
where $B$ is the batch size, $H$ the number of heads, and $T$ the sequence length (number of tokens). 
Masking is applied along the \textbf{Key dimension} (the columns), so that certain tokens cannot be attended to. 
We use two types of masks:

\begin{enumerate}
    \item \textbf{Padding mask.} To enable batching, all sequences are padded, which means appending dummy tokens, typically all zeros, to make every sequence the same length.
    Padding tokens must not affect the model’s output, so we mask them out of the attention computation.  

    \item \textbf{Restricted visibility (local radius).} To study the effect of limiting each token’s visible neighborhood, 
    we compute a restricted attention mask. Allowed interactions are precomputed from the $\mathcal{L}_2$ distances between raw coordinates, and then the same mask is applied to every attention step in the forward pass.  
\end{enumerate}

\paragraph{Key masking mechanism.} 
After computing $QK^\mathsf{T}$, all disallowed positions are replaced with \texttt{-inf}. 
During the row-wise softmax, these entries become zero, ensuring that they cannot contribute, regardless of the values in $V$. 
Consequently, masked tokens never influence the update of valid tokens. 
Query values at masked positions can be arbitrary ``nonsense'' numbers, (some implementations explicitly zero them out after each attention layer for safety and clarity) but they cannot affect non-padded tokens.

\section{Decoder} \label{sec:decoder}
The natural extension of using an encoder to predict forces is to use a decoder to predict energy. While the encoder architecture produces a per-token output \ref{sec:encoders}, the decoder architecture produces individual outputs, like a scalar crystal energy, using a similar Attention/Transformer based architecture. The decoder design we use starts with a stack of encoder layers like in the force-prediction model \ref{sec:encoders}, but instead of the final linear down-scaling, the stack is followed by a decoder head. This head defines a ``dummy'' token, $\mathbf{q}$, which is used to allow the calculations to shrink the output to a constant size. This modification requires us to use a slightly different notation; rather than having $\texttt{Attn}$ as an function of a single variable, we denote it as a function of three variables. Each is used (in order) to provide the conditioning of one of $\mathbf{Q}, \mathbf{K}, \mathbf{V}$.

The Decoder architecture is formulated as:

\begin{align*}
    \mathbf{M} &= \texttt{encoder}(\mathbf{X}); \\
    \mathbf{h}_0 &= \texttt{LN}(\mathbf{q} + \texttt{Attn}(\mathbf{q}, \mathbf{M}, \mathbf{M})); \\
    \mathbf{h}_1 &= \texttt{LN}(\mathbf{h}_0 + \texttt{MLP}(\mathbf{h}_0));\\
    \mathbf{\widehat{y}} &= \mathbf{W}\mathbf{h}_1 + \mathbf{b},
\end{align*}
where the notation follows that used in Section \ref{sec:encoders}, and dropout is applied after $\texttt{Attn}$ and $\texttt{MLP}$. 
Recall that $\mathbf{M} \in \mathbbm{R}^{n \times d_{model}}$, and note that $\mathbf{W} \in \mathbbm{R^{1 \times n}}$. 
Although it is a matrix of shape $\mathbf{q} \in \mathbbm{R}^{(1 \times d_{model})}$ we denote it in lowercase vector form to make clear that it has only one non-trivial dimension. We train both the encoder and decoder layers jointly.

\section{Augmentation Training Experimental Results} \label{sec:fullAugment}
For all experiments included in this section, rotations are performed around the centroid. Each axis has a random translation applied to it independently. Error is calculated as total $\mathcal{L}_2$ error over the testing set. Five trials with random rotations are averaged to calculate the augmented error. 

\begin{table*}[t]
\caption{Training results after 160 epochs with restricted rotations and translations. Reported values are total errors for unaugmented and augmented models.}
\label{tab:rotTransRes}
\begin{ruledtabular}
\begin{tabular}{cccc}
Degree of Rotation & Range of Translation & Unaugmented Error & Augmented Error \\
\hline
0--10   & 0     & 2.753 & 2.726 \\
0--90   & 0     & 13.387 & 3.798 \\
0--360  & 0     & 5.573 & 5.515 \\

0       & 0--0.1 & 1.965 & 1.952 \\
0       & 0--1   & 2.863 & 2.577 \\
0       & 0--10  & 2.659 & 13.788 \\

0--10   & 0--0.1 & 2.687 & 2.740 \\
0--90   & 0--1   & 14.210 & 4.498 \\
0--360  & 0--10  & 6.879 & 13.768 \\
\end{tabular}
\end{ruledtabular}
\end{table*}

\subsection{Preprocessing} 
In some relaxation problems, appropriate preprocessing of the data can eliminate the need for explicit consideration of some symmetries. Structures can be rigidly translated such that they are placed in a pre-determined spot, or coordinates are determined relative to the center of mass. Although more dependent on the specifics of the dataset, it is in some cases even viable to define a canonical ``correct'' orientation and to pre-rotate all structures to match this configuration standard. In systems with the periodic boundary condition (PBC) enforced, we can further shift the location of key features like defect centers to a prescribed location. In our target use case, the high throughput screening of materials defects, candidate structures are all created through the same process by the user. This inherently creates a system in which all inputs share a common alignment by construction. 

Throughout many ML fields, issues with invariances occur \cite{ma2024canonicalization, burgess2024orientation, chang2017clkn, boussot2025impact, pu2019deepdrug3d, martinka2024simple}. In order to enable more expressive architectures, robust preprocessing systems are often introduced. These methods seek to ``register'' inputs into a canonical pose or orientation. Many variations of these methods have been introduced, ranging from simple deterministic shifts and rotations \cite{pu2019deepdrug3d, qi2017pointnet} to advanced neural networks trained to re-orient data \cite{burgess2024orientation, chang2017clkn}. This need has been recognized in atomistic calculations as well \cite{chevrot2011least, martinka2024simple}. In fields conceptually related to MLFF literature, these approaches are often considered standard practice \cite{boussot2025impact, burgess2024orientation, chang2017clkn}, recognizing the need to free architectures from architecturally-costly symmetry constraints. 

\subsection{Tradeoff Reasoning} 
We emphasize that the lack of architectural enforcement of rotational and translational equivariance is a deliberate limitation of the present formulation, and is not intended as a general recommendation for all MLFF architectures. In many settings, explicit equivariance provides a strong inductive bias and can substantially improve data efficiency. However, in the specific context of defect modeling, we find that the additional preprocessing required to approximately respect rigid-motion symmetries is a reasonable tradeoff in exchange for the ability to model long-range interactions without architectural constraints. This suggests a promising research direction in which symmetry handling is partially delegated to preprocessing, allowing architectural capacity to be focused on modeling physically relevant long-range interactions that are otherwise difficult to capture within strictly equivariant formulations.

Alternatively, one may replace absolute coordinates with relative geometric features, following many existing MLFF architectures that operate on pairwise distances and angular information \cite{batatia2022mace, deng2023chgnet, yang2024mattersim, chen2022universal, batzner20223}. In principle, this could be integrated into a Transformer by constructing and slicing a $3 \times n \times n$ tensor of relative distances and angles for attention. However, such formulations incur substantially higher token representations, and we therefore do not pursue them in this work.

\section{Atomic Descriptors} \label{sec:descriptSelect}
\begin{figure*}[t]
    \centering

    \begin{minipage}{0.48\textwidth}
        \centering
        \includegraphics[width=\linewidth]{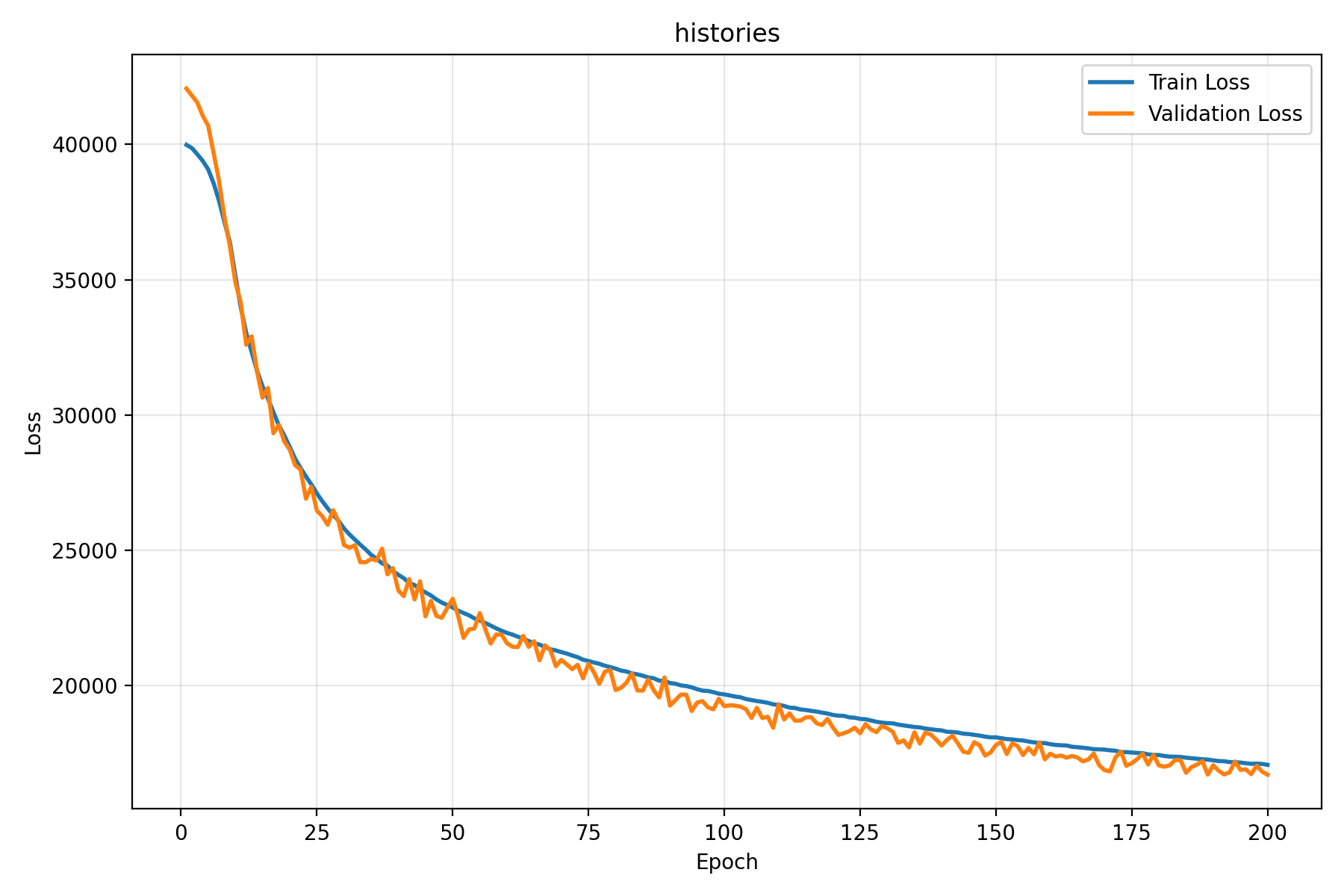}
        \caption{Loss history from training with atomic number only.}
        \label{fig:first}
    \end{minipage}
    \hfill
    \begin{minipage}{0.48\textwidth}
        \centering
        \includegraphics[width=\linewidth]{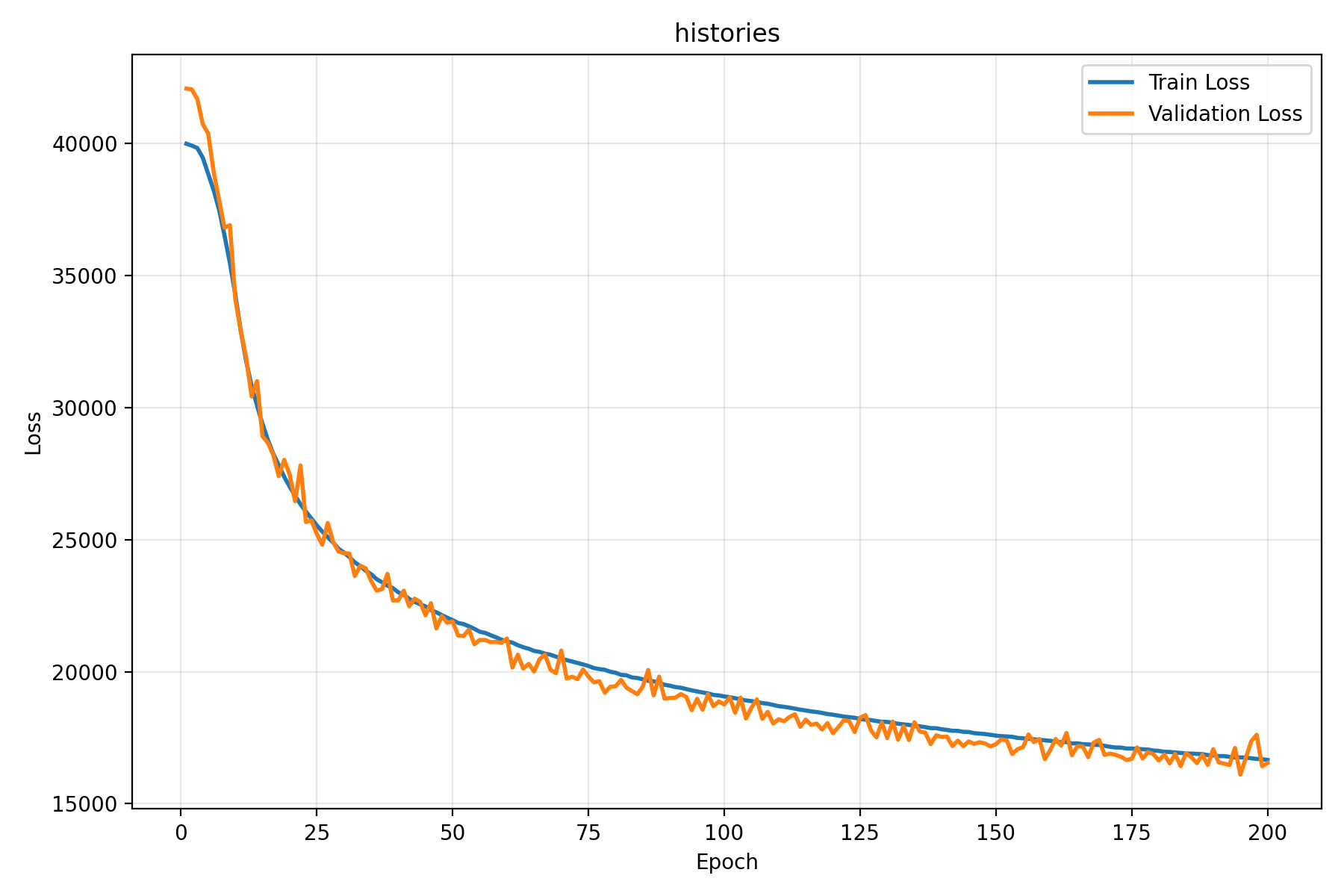}
        \caption{Loss history from training with set of nine atomic descriptors.}
        \label{fig:second}
    \end{minipage}

\end{figure*}

In Figures \ref{fig:first}, \ref{fig:second}, the effect of using the nine atomic descriptors from Section \ref{sec:res} is compared to using just the atomic number to describe atoms. The representations produce extremely similar errors. Finding optimal descriptor sets remains an open problem, and is orthogonal to this work's primary contribution. 

\bibliography{apssamp}

\end{document}